\documentclass[preprint,12pt]{elsarticle}

\usepackage{amssymb}
\usepackage{amsmath}
\usepackage{CJKutf8}
\PassOptionsToPackage{quiet}{fontspec}
\usepackage{amssymb}
\usepackage[textwidth=14.0cm]{geometry}
\usepackage{color} 

\usepackage{makecell}
\usepackage{bm}
\usepackage{booktabs}

\journal{Nuclear Physics B}

\begin{document}
\begin{CJK}{UTF8}{gbsn}

\begin{frontmatter}



\title{Multimodal Fine-grained Reasoning for Post Quality Evaluation}


\author{${Xiaoxu Guo}^{a,b},{Siyan Liang}^{c},{Yachao Cui}^a,{Juxiang Zhou}^{d},{Lei Wang}^a,{{Han Cao}}^{a,{\ast}}$}
\affiliation{organization={School of Computer Science},
            addressline={Shaanxi Normal University}, 
            city={$Xi^{'}an$},
            postcode={710119}, 
            state={Shaanxi Province},
            country={China}}

\affiliation{organization={School of Artificial Intelligence Institute},
	addressline={Beijing Economic Management Vocational College}, 
	city={Beijing},
	postcode={100018}, 
	state={Beijing},
	country={China}}

\affiliation{organization={State Key Laboratory of Information Security, Institute of Information Engineering},
	addressline={Chinese Academy of Sciences}, 
	city={Beijing},
	postcode={100049}, 
	state={Beijing},
	country={China}}

\affiliation{organization={Key Laboratory of Education Informatization for Nationalities},
	addressline={Yunnan Normal University}, 
	city={Kunming},
	postcode={650000}, 
	state={Kunming},
	country={China}}
\begin{abstract}

Accurate assessment of post quality frequently necessitates complex relational reasoning skills that emulate human cognitive processes, thereby requiring the modeling of nuanced relationships. However, existing research on post-quality assessment suffers from the following problems: 1) They are often categorization tasks that rely solely on unimodal data, which inadequately captures information in multimodal contexts and fails to differentiate the quality of students' posts finely. 2) They ignore the noise in the multimodal deep fusion between posts and topics, which may produce misleading information for the model. 3) They do not adequately capture the complex and fine-grained relationships between post and topic, resulting in an inaccurate evaluation, such as relevance and comprehensiveness. Based on the above challenges, the Multimodal Fine-grained Topic-post Relational Reasoning(MFTRR) framework is proposed for modeling fine-grained cues by simulating the human thinking process. It consists of the local-global semantic correlation reasoning module and the multi-level evidential relational reasoning module. Specifically, MFTRR addresses the challenge of unimodal and categorization task limitations by framing post-quality assessment as a ranking task and integrating multimodal data to more effectively distinguish quality differences. To capture the most relevant semantic relationships, the Local-Global Semantic Correlation Reasoning Module enables deep interactions between posts and topics at both local and global scales. It is complemented by a topic-based maximum information fusion mechanism to filter out noise. Furthermore, to model complex and subtle relational reasoning, the Multi-Level Evidential Relational Reasoning Module analyzes topic-post relationships at both macro and micro levels by identifying critical cues and delving into granular relational cues. MFTRR is evaluated using three newly curated multimodal topic-post datasets, in addition to the publicly available Lazada-Home dataset. Experimental results indicate that MFTRR outperforms state-of-the-art baselines, achieving a 9.52$\%$ improvement in the NDCG@3 metric compared to the best text-only method on the Art History course dataset.

\end{abstract}

\begin{graphicalabstract}
\end{graphicalabstract}

\begin{highlights}
\item MFTRR mimics human reasoning for fine-grained post-quality evaluation.

\item Local-global semantic reasoning filters noise for enhanced multi-scales fusion.

\item Evidential reasoning captures complex topic-post relationships at multiple levels.

\item MFTRR outperforms baselines on three new datasets and Lazada-Home.
\end{highlights}


\end{frontmatter}



\section{Introduction}
\footnotetext[1]{*Han Cao is the corresponding author.}  
\footnotetext[2]{E-mail addresses: gxx123@snnu.edu.cn (Xiaoxu Guo), liangsiyuan@iie.ac.cn (Siyan Liang), cuiyachao@snnu.edu.cn (Yachao Cui), zhoujuxiang@ynnu.edu.cn (Juxiang Zhou), wanglei521@snnu.edu.cn (Lei Wang),caohan@snnu.edu.cn (Han Cao)}  
\label{1}

Post-quality assessment has emerged as a significant research direction within the realm of personalized assessment and analytics. With the increasing popularity of digital education platforms and online learning environments, accurately assessing the quality of students' posts plays a crucial role in improving educational outcomes and providing personalized interventions \cite{lalingkar2022models}. In discussion forums, students interact with instructors, generating valuable data regarding the learning process. Effective analysis of this data can help students identify learning problems, adapt their strategies, and enhance their abilities. Furthermore, the data serves as feedback on the instructional effectiveness of teachers. However, much of the existing research on post-quality evaluation primarily focuses on categorization tasks, which limits the ability to analyze subtle changes in student behavior effectively. For example, Anastasios Ntourmas et al. \cite{ntourmas2023classifying} utilized machine learning models to classify posts into three predefined categories. Meanwhile, many studies ignore the noise problem in modal fusion (i.e., the integration of information from different modalities to enhance the representation of information). Xiaohui Tao et al.\cite{tao2023towards} integrated engagement, semantic, and emotional features on a unified scale yet failed to filter out the noise and interfering information during the fusion process. Additionally, most studies fail to capture the complex and nuanced relationships between posts and topics, resulting in inaccurate and incomplete evaluations of post quality. Mohamed A. El-Rashidy et al.\cite{el2024attention} only analyzed the perspective of semantic relationships. As a result, studies on post-quality assessment suffer from fixed assessment scope, difficulty in constructing the complex relationship between posts and topics, and noise problems in modal fusion. In order to solve these problems, it is essential to develop an accurate, multi-relationship clue and fine-grained method (i.e., providing refined and multi-ranging evaluation metrics rather than fixed categories) to assess post quality.
\begin{figure}[htbp]
	\centering
	\begin{tabular}{  c  }
		
		\toprule[2pt]
		\large\makecell[l]{\textbf{Discussion Topic:}}   \\ 
		
		\makecell[l]{矩阵高次幂运算的思考} \\ 
        
		\begin{minipage}[b]{0.7\columnwidth}
			\centering
            
			\raisebox{-.2\height}{\includegraphics[width=\linewidth]{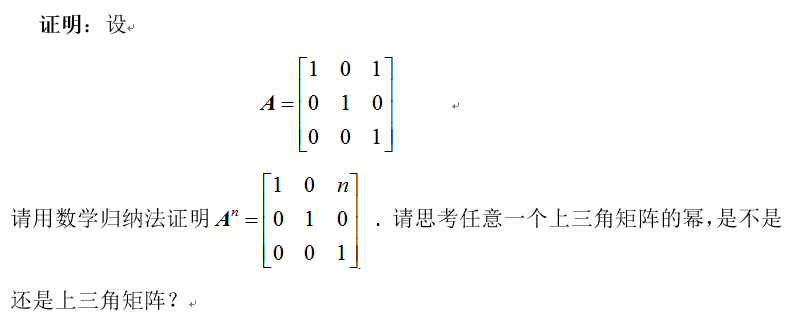}}
		\end{minipage}
		\\ \hline
		\large\makecell[l]{\textbf{post1(score:3):}}   \\
		\makecell[l]{证明如下}  \\
		\begin{minipage}[b]{0.5\columnwidth}
			\centering
   
			\raisebox{-.2\height}{\includegraphics[width=\linewidth]{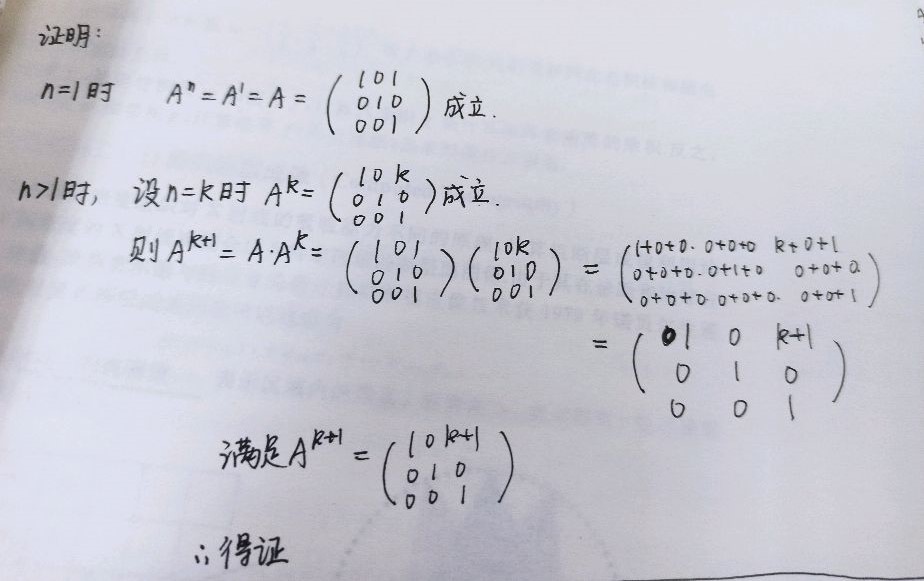}}
		\end{minipage}
		
		
		\\ \hline
		\large\makecell[l]{\textbf{post2(score:4):}}   \\
		
		\begin{minipage}[b]{0.5\columnwidth}
			\centering
			\raisebox{-.5\height}{\includegraphics[width=\linewidth]{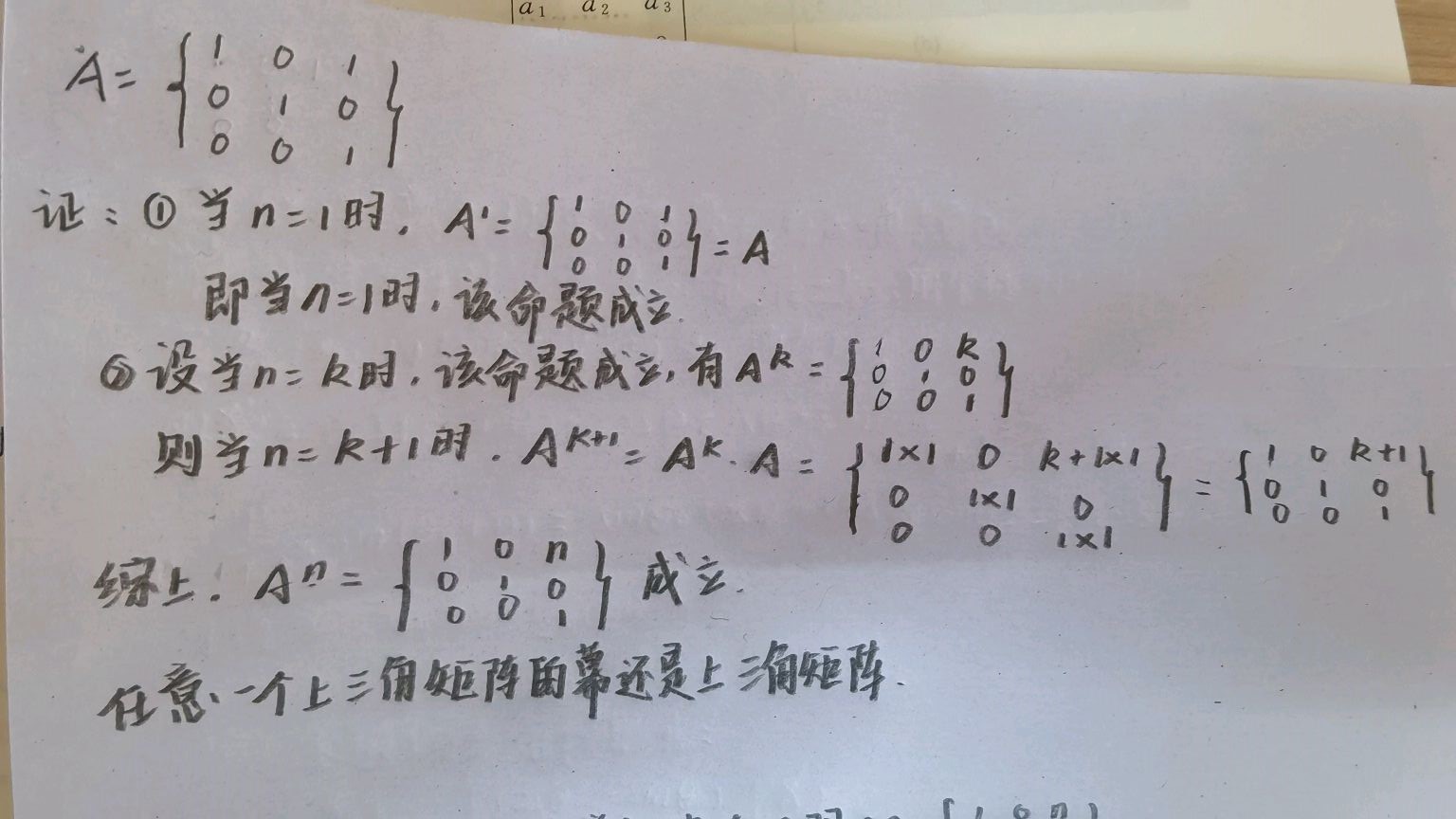}}
		\end{minipage}
		
		\\
		\bottomrule[2pt]
	\end{tabular}
	\caption{Multimode post examples under the topic "Thinking about high-order matrix Exponentiation operation". The text of Post 1:   “The proof is as follows” . The answer of Post 2 is displayed in the form of pictures. Pure text can correctly predict the quality of the corresponding discussion topic. (\textbf{Top}) The discussion topic is translated as follows, "Thinking about high-order matrix Exponentiation operation".}
\end{figure}

\begin{figure}[htbp]
	\centering
	\begin{tabular}{  c  }
		\toprule[2pt]
		\large\makecell[l]{\textbf{Discussion Topic:}}   \\ 
		\makecell[l]{佛教造像的基本常识，其中包含的内容较多，欢迎大家分} \\  
        \makecell[l]{享佛教造像的知识，并上传相关图片说明。} \\  
		\begin{minipage}[b]{0.45\columnwidth}
			\centering
			\raisebox{-.5\height}{\includegraphics[width=\linewidth]{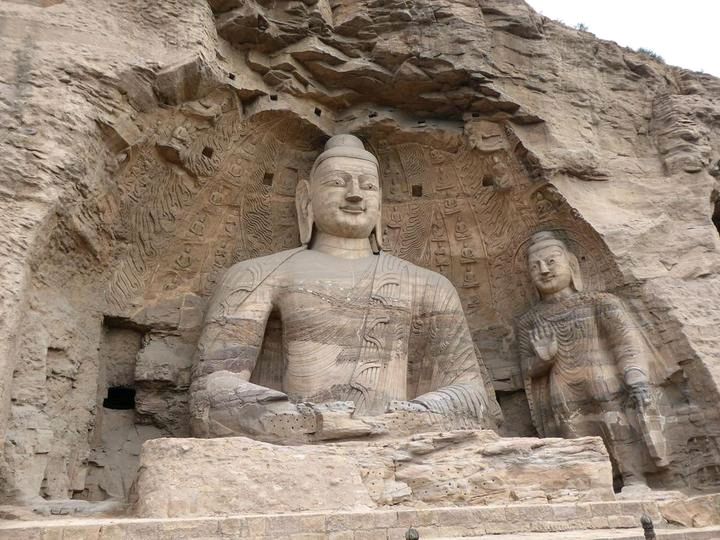}}
		\end{minipage}
       
		\\ \hline
		\large\makecell[l]{\textbf{post1:}}   \\
		\makecell[l]{佛教造像始自印度时代的犍陀罗，在印度各地不断完善，}  \\
		\makecell[l]{之后才传向东南亚各国和中国。}  \\
		\begin{minipage}[b]{0.3\columnwidth}
			\centering
			\raisebox{-.5\height}{\includegraphics[width=\linewidth]{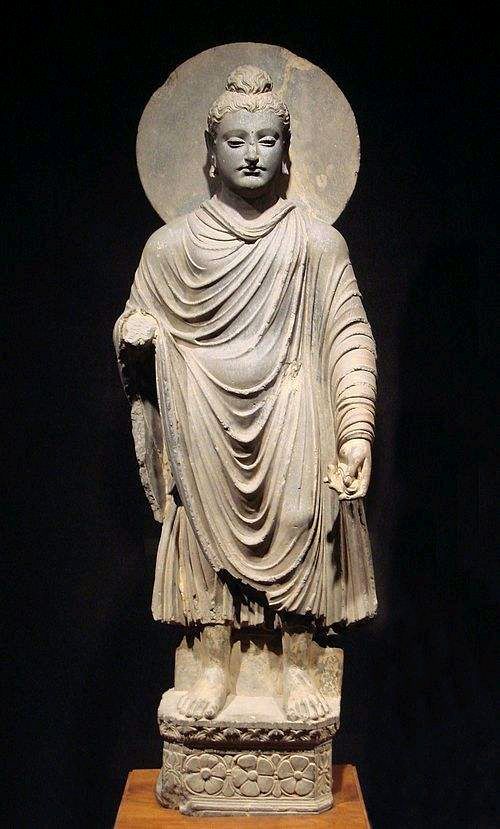}}
		\end{minipage}
		\\
		\bottomrule[2pt]
	\end{tabular}
	\caption{Multimode post examples under the topic "Buddhist statue". Using text alone does not fully answer the question of the discussion topic, and the attached images answer the second question. (\textbf{Top}) The text content of the discussion topic: "There is a lot more basic knowledge about Buddhist statues.  Welcome to share the knowledge about Buddhist statues and upload related images with descriptions.“; (\textbf{Bottom}) The text content of post 1: "Buddhist statues began in Gandhara during the Indian era and were perfected throughout India before spreading to the countries of Southeast Asia and China".}
               
\end{figure}

\begin{figure}[htbp]
	\centering
	\begin{tabular}{  c  }
		\toprule[2pt]
		\large\makecell[l]{\textbf{Discussion Topic:}}   \\ 
		\makecell[l]{结合图讨论：哪些因素造成了同时期古代希腊与古代中国形} \\  
        \makecell[l]{成了不同的政治制度？有何影响？} \\  
		\begin{minipage}[b]{0.45\columnwidth}
			\centering
			\raisebox{-.5\height}{\includegraphics[width=\linewidth]{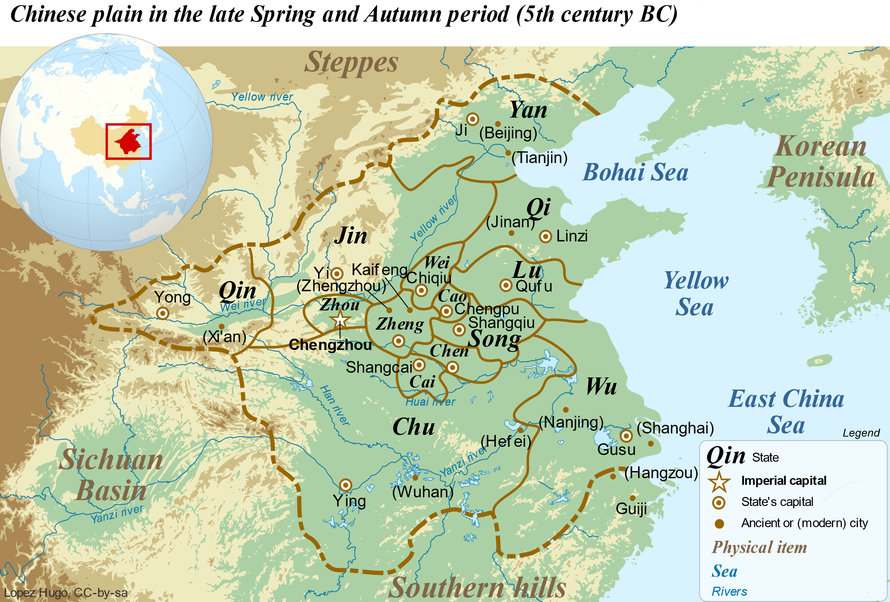}}
		\end{minipage}
       
		\\ \hline
		\large\makecell[l]{\textbf{post1:}}   \\
		\makecell[l]{最主要因素是地理结构和海陆位置}  \\     
		\\
		\bottomrule[2pt]
	\end{tabular}
	\caption{Multimode post examples under the topic "Greece and China". Analyzing posts in relation to the topic using only a single dimension and perspective does not accurately assess post quality. It takes multi-levels and multi-perspectives of analysis to realize that the post does not answer both questions, but only the first. (\textbf{Top}) The text content of the discussion topic: "Discuss with the diagram: what factors led to the development of different political systems in Ancient Greece and Ancient China during the same period? What were the implications?“; (\textbf{Bottom}) The text content of post 1: "The main factors are geographical structure and land-sea location".}
               
\end{figure}

It is becoming increasingly common to use various forms of data to answer questions on online education platforms. Relying exclusively on textual data to evaluate the quality of posts limits understanding of posts and topics discussed. As shown in Figure 1, some students responded to the discussion topic in the form of pictures. However, assessments based solely on textual information can result in inaccurate evaluations. Figure 2 presents a discussion topic that encouraged students to share their knowledge about Buddhist statues and to upload relevant images. Students provided comprehensive responses that combined both images and text. When analyzing post quality based solely on textual data, it appears that students engaged with the first question but neglected the second question regarding the upload of relevant images. One possible hypothesis that arises from these examples is that assessing post quality solely through textual data analysis is inadequate and that combining text and additional images is more conducive to accurately assessing post quality. Figure 3 illustrates that the discussion topic posed two questions, yet the post did not answer both questions but only the first. Analyzing the relationship between posts and topics from a single-dimensional perspective fails to accurately assess the post quality. Therefore, it is crucial to fully capture the complex relationship between posts and topics from multiple levels and perspectives. Therefore, it is crucial to fully capture the complex relationship between posts and topics from multiple levels and perspectives. This multidimensional understanding is not only essential for improving educational engagement analysis but also serves as a foundation for addressing potential safety concerns in multimodal content environments~\cite{liu2025natural,liang2024unlearning,kuang2024adversarial,lu2025adversarial,wang2025manipulating,ying2025reasoning,liang2023badclip,liang2024revisiting}.

Due to the extensive data resources available in discussion forums, an increasing number of researchers are focusing on the evaluation of post quality within these platforms. Earlier studies used traditional machine learning and statistical methods to analyze post-quality. For example, A. Chanaa, et al.\cite{chanaa2022sentiment} used machine learning methods to detect sentiment states in Massive Open Online Courses (MOOCs). Almatrafi et al. \cite{almatrafi2018needle} analyzed post quality by the machine learning model. However, the limitations of traditional machine learning, such as being time-consuming, sparse, and having unbalanced samples, significantly impede the development of evaluating post quality\cite{moon2024applying}. In recent years, deep learning has emerged as the predominant research direction in post-quality analysis, primarily due to its ability for automatic feature extraction. For example, Sannyuya Liu et al.\cite{liu2022automated} constructed a deep neural network architecture BER-CNN to analyze the quality of posts. Purnachary et al. \cite{munigadiapa2022mooc} proposed an LSTM variant network to analyze the quality of posts. Nevertheless, several challenges remain in the  existing research on post-quality assessment: (1)Most current studies primarily concentrate on categorical tasks using solely textual data, which limits their ability to accurately and fine-grained assess post-quality and detect subtle gaps in students' performance. (2) An accurate and comprehensive assessment of a post necessitates a multi-dimensional and multi-level analysis to fully capture the complex relationship between the post and its topic. (3) Many studies lack a fine-grained fusion of multimodalities at different scales, which is essential for obtaining adequate semantic information, and they often overlook the problem of noise in the fusion process.

To address the above challenges, the Multimodal Fine-Grained Topic-Post Relational Reasoning (MFTRR) framework is proposed. The framework utilizes multimodal data to finely assess post quality by reconstructing human thought from multi-scale and multi-level. It consists of the Local-Global Semantic Correlation Reasoning module and the Multi-Level Evidence relational Reasoning module. The Local-Global Semantic Relevance Reasoning module determines whether a post adequately covers all aspects of the discussion topic by analyzing the semantic relationship between the post and the topic at both global and local scales. The Multi-Level Evidence Relationship Reasoning module assesses whether the post provides an accurate and fine-grained response to the discussion topic's questions by analyzing the relationship between the post and the topic at both macro and micro levels. Specifically, the Local-Global Semantic Correlation Reasoning module is introduced to effectively extract maximum semantic relational information through fusion at different ranges of topics and posts and noise filtering. Meanwhile, the Multi-Level Evidential Relational Reasoning module is designed to capture complex and fine-grained relationships between posts and topics through Topic-Post Significant Information Evidence Reasoning and Topic-Post Internal Logic Relationship Evidence Reasoning from both macro and micro levels. This module will provide robust evidence for assessing topic-post quality. Furthermore, to accurately capture the differences in quality among various posts under the same topic, the post-quality assessment is set as a ranking problem, and pairwise ranking objectives are used to optimize the model. 

The main contributions are summarized as follows:

(1)This paper proposes the multi-modal fine-grained topic-post relational reasoning （MFTRR）method, which is designed to mimic the human thought process. The method leverages the local-global semantic relation reasoning module and the multi-level evidence relation reasoning module to achieve a refined evaluation of post quality.

(2)This paper proposes a local-global semantic reasoning module that facilitates deep fusion and interaction between posts and topics at both local and global scales. Meanwhile, the topic-based global fusion mechanism is employed to filter out noise, thereby maximizing the extraction of semantically relevant information.

(3)This paper proposes the multi-level evidential relational reasoning module captures subtle relationships through topic-post significant information evidence reasoning and topic-post internal logic relationship evidence reasoning. The former retrieves significant information oriented towards relationship graphs at the macro level, while the latter constructs relationships between internal topics and posts at the micro level.

(4) Comprehensive experiments conducted on three newly collected datasets, along with the public Lazada-Home dataset, demonstrate that MFTRR outperforms various baseline methods. Notably, it exceeds the performance of the best text-only method by 9.52$\%$ in terms of the NDCG@3 metric on the Art History course dataset.

\section{Related Work}
\subsection{Multimodal Fusion}
\label{subsec1}

Multimodal data, by integrating diverse information, offers a powerful tool for holistic comprehension and analysis of complex issues, thereby attracting the interest of educational researchers \cite{ouhaichi2023research}. Currently, multimodal fusion technology has been applied to sentiment analysis\cite{wang2024multimodal}\cite{zhao2024multimodal}, learning engagement evaluation \cite{li2024multimodal}\cite{liu2024profiling}, and  behavioral analysis of both teachers\cite{luo2022three}\cite{wu2020recognition} and students\cite{bhattacharjee2022multi} in the field of education. However, research on multimodal data in discussion forums remains scarce, particularly concerning the deep fusion of multimodal data and noise filtering. Multimodal fusion techniques can be categorized into early fusion, late fusion, and hybrid fusion based on the fusion stages. Early fusion entails extracting different modal features and concatenating them without any interaction or overlap. Zhi Liu et al.\cite{liu2023dual} extracted the explicit features and hidden features from discussion forums to jointly analyze students' engagement. Later fusion, which involves using different classifiers to extract features from various modal data before fusing them together, can be time-consuming. Xiaohui Tao et al. \cite{tao2023towards}analyzed students' emotional states by separately extracting engagement, semantics, and emotion features from discussions prior performing fusion. Hybrid fusion utilizes multiple fusion methods in the process of multimodal fusion \cite{chango2021improving}\cite{song2023mifm}. When evaluating the quality of a post, teachers consider both the post text and image in relation to the topic's text and image, as well as the overall coherence of the post with the topic. This evaluation process encompasses multiple interactions between different modalities and interactions between the post and the topic based on the text and images as a whole. However, current research lacks effective noise filtering in the multimodal fusion process, as well as multi-scale and fine-grained fusion techniques.

\begin{table}[htbp]
    \centering
    \begin{tabular}{|>{\tiny}m{1.2cm}|>{\tiny}m{7cm}|>{\tiny}m{4cm}|}
        \hline
        \textbf{Method} & \textbf{Strengths} & \textbf{weaknesses} \\
        \hline
        B-LIWC-UDA\cite{liu2023dual} & 
        1.Integrate semantic and implicit cognitive features from textual information for comprehensive representation.\newline
        2.Effectively address the issue of data scarcity through semi-supervised learning. & 
        1.The model's performance relies on the quality of the LIWC dictionary.\newline
        2.Only text information from the discussion forum is utilized. \newline
        3.The dependency relationships between modalities are not constructed.\\
        \hline
        ESSE\cite{tao2023towards} & 
        1.Feature fusion combining engagement, semantics, and sentiment characteristics.\newline
        2.Ensemble multiple machine learning models for prediction. & 
        1.Only extract features from textual information.\newline
        2.Lack of deep interactions between modalities. \\
        \hline
        MKGE\cite{bhattacharjee2022multi} & 
        1.Combine learning activity features with post textual information.\newline
        2.Interactions have been conducted between modalities. & 
        1.Only the textual information of posts was used, even when the posts contained multimodal data.\newline
        2.Modal fusion was only conducted within a single scale.\newline
        3.The influence of noise within individual modalities on modal fusion is ignored. \\
        \hline
        TDLIM\cite{luo2022three} & 
        1.The model analyzes by integrating multimodal data information.\newline
        2.The three modalities are fused through weighted hierarchical fusion and integrated through decision-level fusion. & 
        1.Ignore the issue of noise in multimodal fusion.\newline
        2.Conduct modal fusion within a single scale. \\
        \hline
    \end{tabular}
    \caption{Summarize the main methods of modal fusion used in post-analysis and analyze their strengths and weaknesses.}
    \label{tab:methods1}
\end{table}

\subsection{Cross-modal Retrieval}

\begin{table}[htbp]
    \centering
    
    \begin{tabular}{|>{\tiny}m{1.2cm}|>{\tiny}m{7cm}|>{\tiny}m{4cm}|}
        \hline
        \textbf{Method} & \textbf{Strengths} & \textbf{weaknesses} \\
        \hline
        MMCA-CMR\cite{tian2020deep} & 
        1.Through effective nonlinear mappings to a common representation space, the embeddings simultaneously preserve both the original feature information and semantic information within this shared space. This aids in reducing the heterogeneity gap between different modalities of data.\newline
        2.By minimizing both the discriminative loss in the common representation space and the prediction loss within this shared space, an effective metric for measuring semantic similarity between multimodal data is achieved. & 
        1.Models rely on high-quality data.\newline
        2.The impact of noise on the model's generalization ability is ignored. \\
        \hline
        DRSL\cite{wang2021drsl} & 
        1.Effectively learning pairwise similarities to bridge the heterogeneous gap across different modalities.\newline
        2.Adequately addressing issues of information imbalance and disparity among different modalities.\newline
        3.Capturing implicit nonlinear relational distance through learning a relational network module. & 
        1.Cross-modal feature fusion is only performed at a single scale, lacking exploration of multi-scale integration.\newline
        2.The computational cost is high. \\
        \hline
        COTS\cite{lu2022cots} & 
        1.Enhancing model performance through three levels of cross-modal feature interaction.\newline
        2.Pre-training the model on large-scale data to endow it with good generalization ability. & 
        1.Model pre-training relies on large-scale data.\newline
        2.High computational resource requirements. \\
        \hline
        ASCSH\cite{meng2020asymmetric} & 
        1.The model decomposes the mapping matrix into consistency and modality-specific components, enabling more accurate discovery of intrinsic semantic correlations between modalities.\newline
        2.The model leverages pairwise similarities and semantic labels jointly to guide the learning of hash codes, enhancing their discriminative ability. & 
        1.High computational resource requirements.\newline
        2.The model relies on a large amount of high-quality training data. \\
        \hline
        CLIP\cite{radford2021learning} & 
        1.The model possesses good zero-shot generalization capability.\newline
        2.Through large-scale pre-training, CLIP can effectively measure the semantic similarity between images and text. & 
        1.CLIP exhibits poor performance in fine-grained analysis.\newline
        2.High computational resource requirements. \\
        \hline
    \end{tabular}
    \caption{Summarize the strengths and weaknesses of key approaches in cross-modal retrieval.}
    \label{tab:methods2}
\end{table}

Cross-modal retrieval aims to establish semantic links across different modalities by addressing their inherent heterogeneity. It can be categorized into real-valued representation learning and binary-coded representation learning \cite{gu2018look}. Real-valued methods focus on projecting features into a shared space, where the closest vector is queried to facilitate retrieval. Some methods use typical correlation analysis (CCA) \cite{abdi2018canonical} to learn the common representation by maximizing the pairwise correlation between two modalities. However, addressing nonlinear correlations between features poses significant challenges. Deep learning-based methods \cite{tian2020deep} have demonstrated promise in effectively managing these nonlinear correlations across different modalities. Wang et al. \cite{wang2021drsl} proposed a deep hybrid framework model that obtains a common representation by directly learning natural pairwise similarities. Additionally, Haoyu Lu et al. \cite{lu2022cots} introduced a collaborative two-stream visual language pre-training model that incorporates token-level interactions and task-level interactions to enhance instance interactions for a more effective representation space. The binary representation learning approach projects cross-modal data into a common Hamming space, assigning similar hash codes to similar cross-modal content. Min Meng et al. \cite{meng2020asymmetric} proposed an Asymmetric Supervised Consistent and Specific Hashing approach, which decomposes the mapping matrix into consistent and specific modal matrices to fully explore the correlation between different modalities, enhancing the discriminative ability of hash codes. Furthermore, Kaiyi Luo et al. \cite{luo2023adaptive} presented an Adaptive Marginalized Semantic Hashing (AMSH) method that enhances discrimination between latent representations and hash codes using adaptive boundaries, thereby improving the differentiation and robustness of the hash code. Li Li et al.\cite{li2024robust} proposed a robust online hashing method that incorporates label semantic enhancement. This method leverages low-rank and sparse constraints to ensure clean labels and establishes relationships between new and old data through similarity measures, resulting in better performance. In semantic similarity analysis, some studies leverage substantial amounts of web data to understand semantic relationships, as exemplified by ALIGN \cite{jia2021scaling} and CLIP \cite{radford2021learning}. Nonetheless, these methods impose high computational demands, making them impractical for real-world applications, and they do not offer a comprehensive analysis. Therefore, it is essential to explore computationally efficient approaches that consider multi-levels in addressing cross-modal interactions.

\subsection{Reviewing Discussion Forum}

\begin{table}[htbp]
    \centering
    \begin{tabular}{|>{\tiny}m{1.2cm}|>{\tiny}m{5.5cm}|>{\tiny}m{5.5cm}|}
        \hline
        \textbf{Method} & \textbf{Strengths} & \textbf{weaknesses} \\
        \hline
        MLAD\cite{khan2020machine} & 
        1.the model combines lexical features, non-lexical features and new semantic features to improve accuracy.\newline
         & 
        1.Only the text information from the discussion forum is utilized.\newline
        2.Manual feature extraction is required, which is a time-consuming process. \newline
        3.The evaluation categories are fixed.\\
        \hline
        CMSS\cite{ntourmas2023classifying} & 
        1.The model leverages the semantic similarity between dynamically created corpora and forum posts in the MOOC environment as training features.\newline
         & 
        1.Only text information from the discussion section is utilized.\newline
        2.Features need to be manually extracted, which is a time-consuming process.\newline
        3.The evaluation categories are fixed. \\
        
        \hline
        MOOC-BERT\cite{liu2023mooc} & 
        1.Features are automatically extracted, eliminating the need for feature engineering.\newline
        2.The model outperforms traditional machine learning models in terms of performance. & 
        1.Only discussion forum text information is utilized.\newline
        2.The model requires pre-training.\newline
        3.The evaluation categories are fixed. \\
        \hline
        AHRN\cite{capuano2021attention} & 
        1.The model is capable of learning textual context information and extracting key information.\newline
        2.Features are automatically extracted, eliminating the need for feature engineering. & 
        1.Only text information from the discussion forum is used.\newline
        2.The evaluation categories are predetermined.\newline
        3.It lacks the integration of information from multiple scales.  \\
        \hline
    \end{tabular}
    
    \caption{Summarise the main methods for researching discussion forums, their strengths and weaknesses.}
    \label{tab:methods3}
\end{table}

In discussion forum studies, which primarily involve textual information, methods are typically categorized into three groups: statistical methods, machine learning methods, and deep learning methods\cite{zhang2024bert}. Statistical methods involve the quantitative analysis of relationships between posts in discussion forums\cite{wise2017mining}. This process can be time-consuming and requires experienced experts for accurate quantitative analysis. Machine learning methods leverage domain-specific knowledge to manually extract features, which are then input into traditional classifiers for prediction, such as Support Vector Machines (SVM) and Random Forest. Atif Khan et al. \cite{khan2020machine}introduced LinearSVC, a variation of SVM, to assess the effectiveness of replies in discussion forums. Anastasios Ntourmas et al. \cite{ntourmas2023classifying} extracted semantic similarity phase features of discussion boards and fed them into a machine learning model to classify discussions into three categories: course logistics, content-related, and no action required. Almatrafi et al.\cite{almatrafi2018needle} addressed the issue of information overload in discussion forums by employing a machine learning model, including Naïve Bayes, Support Vector Machines, Random Forests, and AdaBoost, to identify urgent posts. However, the process of manual feature extraction remains both time-consuming and labor-intensive. Several recent studies have utilized deep neural networks to automatically extract post features for analysis. Munigadiapa, Purnachary et al.\cite{munigadiapa2022mooc} conducted student sentiment analysis on posts using Long Short-Term Memory Architecture (LSTM) and Axe Hyperparametric Tuner. Zhi Liu et al. \cite{liu2023mooc} introduced MOOC-BERT, a novel variant of the Transformer Bidirectional Encoder Representation Model designed to recognize students' cognitive states. Nicola Capuano et al. \cite{capuano2021attention} proposed a hierarchical recurrent neural network that utilizes an attention mechanism to capture contextual information through a Bidirectional Recurrent Neural Network (BiRNN) and subsequently extract key information via the attention mechanism. Subsequently, some works explored the relationship between posts and learning behaviors by analyzing discussion board data\cite{zou2021exploring}\cite{jin2025sentiment}. However, most of these methods only consider textual information from discussion forums while neglecting other modalities. Furthermore, they tend to analyze data at a singular level and rely on fixed evaluation categories. To address these limitations, the task set as a rank task and a deep neural network architecture is proposed that integrates multimodal data, achieving a comprehensive assessment of post quality through multi-scale fusion and multilevel analysis of the relationship between topics and posts.

\section{Methodology}

\begin{figure}[htbp]
	\centering
    \includegraphics[scale=0.5]{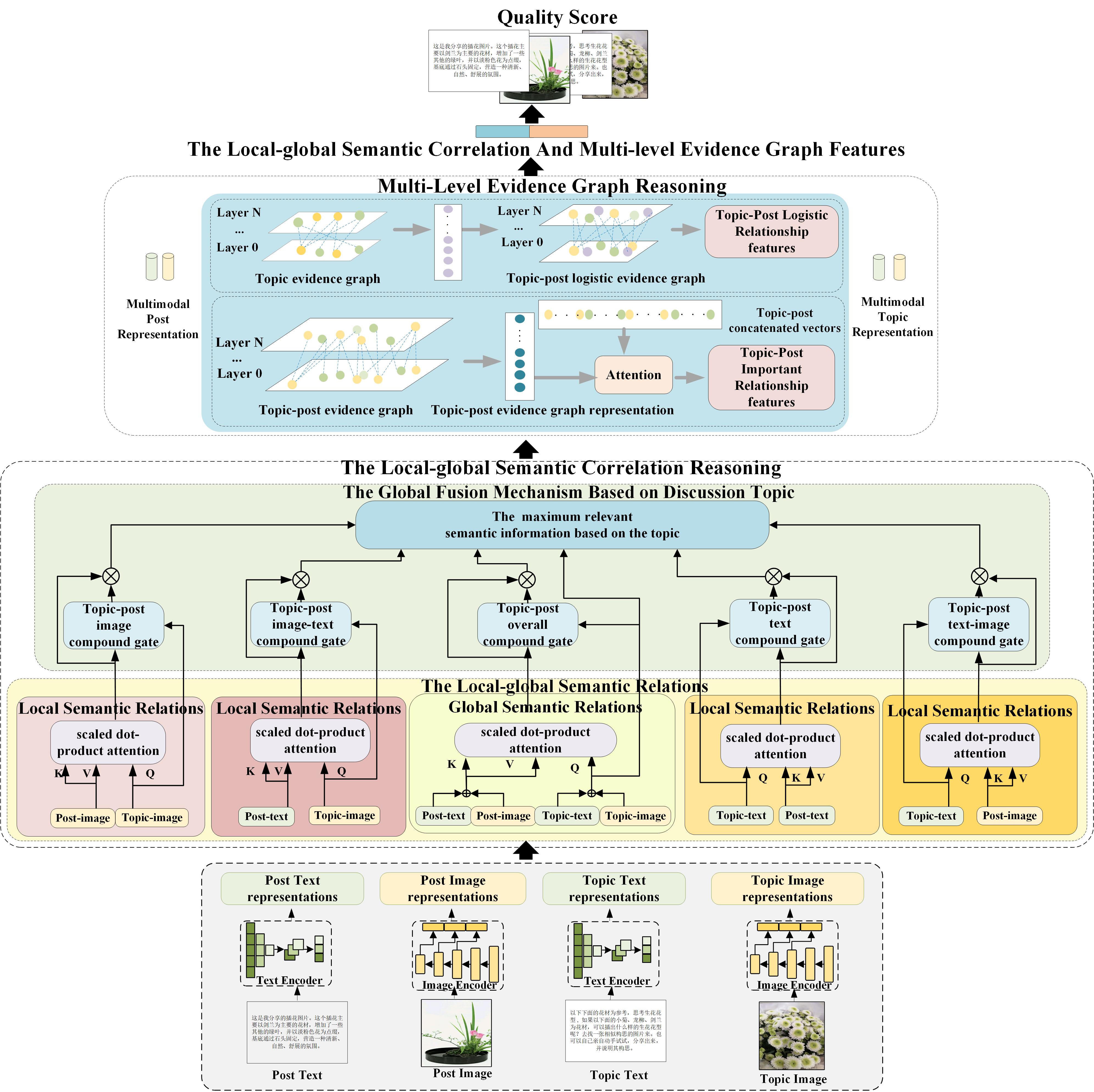}
    
	\caption{Model overview of the end-to-end MFTRR method, which consists of two components: the local-global semantic correlation reasoning module and the multi-level evidential relational reasoning Module.}
	\label{fig:1}
\end{figure}

To explore the intricate relationships and subtle cues between posts and topics, the Multimodal Fine-Grained Topic-Post Relationship Reasoning (MFTRR) framework is proposed. This framework mimics the human thought process through the Local-Global Semantic Correlation Reasoning Module and the Multi-Level Evidential Relational Reasoning Module, assessing the quality of posts in a fine-grained manner. The Local-Global Semantic Correlation Reasoning Module analyzes the semantic relationships between a post and a topic within different scales to determine whether the posts cover all aspects of the discussion topic. Meanwhile, the Multi-Level Evidential Relational Reasoning Module analyzes the hierarchical relationships between post and topic from different levels to determine whether the post accurately respond to the questions posed by the discussion topic. Specifically, the Local-Global Semantic Correlation Reasoning Module extracts semantic correlation information by performing multimodal interactions between posts and topics at both local and global scales. Subsequently, noise is filtered by a global fusion mechanism based on discussion topics to obtain valuable semantic correlation information. The Multi-Level Evidential Relational Reasoning Module captures subtle relationships across multiple levels through topic-post significant information evidence reasoning and topic-post internal logic relationship evidence reasoning at both macro and micro levels. The topic-post significant information evidence reasoning utilizes a retrieval attention mechanism that draws from the post-topic overall relational evidence graph and the post-topic representation to obtain relationship graph-oriented significant information at the macro level. Meanwhile, the topic-post internal logic relationship evidence reasoning constructs a relationship evidence graph by combining the overall internal relationship of the topic with the post to obtain the subtle relationship features between the topic and the post at a micro level. Figure \ref{fig:1} illustrates the overall structure of the current multimodal fine-grained reasoning framework. In the following section, each component of the multimodal fine-grained inference framework is described in detail.

\subsection{Problem Definition}


To uncover the subtle differences between posts, the topic-post quality assessment problem is formulated as a ranking task. This methodology helps instructors capture student learning at a holistic level while also detecting nuanced differences at the individual level. Specifically, a discussion topic $T_i$ consists of its related information (text description and associated image) and a series of posts around this topic $P_i=\{p_{i,1},......,p_{i,N}\}$, where $N$ denotes the number of posts for a discussion topic $T_i$. Each post has a scalar label $S_{i,j}=\{0,......,S\}$, indicating the quality score of the post. The ground-truth ranking of $P_i$ relies on the topic-post quality score, arranged in descending order. The topic-post quality score is determined by analyzing whether posts $p_{i,j}$ comprehensively and accurately answer the question of discussion topic $T_i$. The topic-post quality score $\hat{S}_{i,j}$ is defined as:
\begin{equation}
	\hat{S}_{i,j}=F\{T_i,p_{i,j}\}  \label{eq:commutative1}
\end{equation}
Where $F$ is the post quality score prediction function and $<T_i,p_{i,j}>$ denotes the discussion topic-post pair as input. The discussion topic $T_i$ includes a text description $D_t$ and an associated image $I_t$, and the post $p_{i,j}$ consists of the student-posted text $D_p$ and the image $I_p$.

\subsection{Feature Representation}
\subsubsection{Text Representation}

Convolutional Neural Networks (CNNs) have gained popularity in natural language processing due to their effectiveness and parameter efficiency\cite{dai2018convolutional}. In this paper, the convolutional neural network is used to extract features from text. Specifically, a text (topic text $D_t$ or post text $D_p$) consists of $l_D$ text tokens $\{w_1,......,w_{l_D}\},w_i\in{\mathbb{R}^d}$. Each word is initially converted into a word vector via a word embedding layer. Subsequently, these word vectors are dispatched to multiple one-dimensional CNNs for the extraction of $C-gram$ representations. The conversion process is represented as:
\begin{equation}
	 \mathbf{W}^C=CNN^C(embedding\{w_1,......,w_{l_D}\})
	  \label{eq:commutative2}
\end{equation}
Where $C\in{\{1,......C_{max}\}}$ denotes the size of the kernel. Kernels of different sizes are used to perform the convolution operation. All $C-gram$ representations are subsequently combined to form the final text representation, denoted as $\mathbf{W}=[\mathbf{W}^1,......,\mathbf{W}^{C_{max}}]$. $\mathbf{W}^C\in{\mathbb{R}^{l_D\times{d_D}}}$  is the $C-gram$ representation. $W_t$ and $W_p$ are used to denote the textual feature representations of a discussion topic and a post, respectively.

\subsubsection{Image Representation}

To enrich the semantic information of visual features, the pre-trained backbone network CSPDarkNet \cite{wang2020cspnet} is used to extract multi-scale visual features. Three multi-scale outputs, denoted as $\mathbf{v}_1$, $\mathbf{v}_2$, and $\mathbf{v}_3$, are obtained from the CSPDarkNet backbone network. These three scale representations are then concatenated to form the final multi-scale image representation V. The dimensions of $\mathbf{v}_1$, $\mathbf{v}_2$, and $\mathbf{v}_3$ are $d_{s1}×d_{s1}×d_{1}$, $d_{s2}×d_{s2}×d_{2}$, and $d_{s3}×d_{s3}×d_{3}$, respectively, where $d_{si}×d_{si}$ represents the image scale and $d_i$ represents the feature map, with $i = {1,2,3}$.
\begin{equation}
	\mathbf{v}_i=CSPDarkNetvisual(\{{I}\}), {I}\in{\{{I_t},{I_p}\}}
	\label{eq:commutative3}
\end{equation}
\begin{equation}
	\mathbf{V}=(\mathbf{v}_1\oplus{ \mathbf{v}_2}\oplus{\mathbf{v}_3})
	\label{eq:commutative3_1}
\end{equation}
Where $\mathbf{V}\in{\mathbb{R}^{l_I\times{d_I}}}$ denotes the visual representation, $l_I$ denotes the number of visual features , and $d_I$ denotes the dimension. $\oplus$ denotes the concatenation of vectors. The image representations of the discussion topic and post are denoted by $V_t$ and $V_p$, respectively.

\subsection{The Local-global Semantic Correlation Reasoning}

When reviewing whether a student's post comprehensively answers the topic question, teachers will analyze the semantic relationship between the post and the topic from multiple scopes. Firstly, they examine the semantic relationship between the post and the topic at the textual or imagery level. Then, the semantic relationship between the post and the topic from a holistic perspective is analyzed to determine whether it fully covers all aspects of the discussion topic. Based on the above idea, a local-global semantic correlation reasoning module is designed to simulate this process. Since each modality originates from a distinct embedding space, conducting a direct inter-modal correlation analysis may not retain the maximum relevant information but rather amplify the presence of noise. Therefore, the text representation $W$ and the image representation $V$ are initially projected into a $d_c$-dimensional common latent space: 

\begin{equation}
	\mathbf{H^W}=Tanh(\mathbf{A}_1\mathbf{W}+\mathbf{b}_1)
	\label{eq:commutative4}
\end{equation}
\begin{equation}
	\mathbf{H^V}=Tanh(\mathbf{A}_2\mathbf{V}+\mathbf{b}_2)
	\label{eq:commutative5}
\end{equation}
Where $\mathbf{H^W}\in{\mathbb{R}^{l_D\times{d_c}}}$ and $\mathbf{H^V}\in{\mathbb{R}^{l_I\times{d_c}}}$ denote the text representation and image representation in latent space, respectively.

\subsubsection{The Local-global Semantic Relations}
To fully grasp the semantic relationship between posts and topics, the semantic relationships are analyzed on two scales: local and global. Initially, the text or image of the post and the text or image of the topic are semantically analyzed separately. Subsequently, the semantic relationship between the post and the topic as a whole based on the text and image is analyzed. The traditional scalar dot product attention mechanism is employed for its ability to quickly and efficiently analyze semantic relationships, as well as the advantage provided by the scaling factor that facilitates optimization \cite{kumar2022did}. This paper uses traditional scaled dot-product attention to analyze the semantic relationship between posts and topics. The specific process is shown below:

\begin{equation}
	\mathbf{M}_{W-W}=softmax( \frac{\mathbf{H}_{p}^W({\mathbf{H}_{t}^W}){}^T}{\sqrt{d}})\mathbf{H}_{t}^W
	\label{eq:commutative6}
\end{equation}
\begin{equation}
	\mathbf{M}_{W-V}=softmax( \frac{\mathbf{H}_{p}^W({\mathbf{H}_{t}^V}){}^T}{\sqrt{d}})\mathbf{H}_{t}^V
	\label{eq:commutative7}
\end{equation}
\begin{equation}
	\mathbf{M}_{V-W}=softmax( \frac{\mathbf{H}_{p}^V({\mathbf{H}_{t}^W}){}^T}{\sqrt{d}})\mathbf{H}_{t}^W
	\label{eq:commutative8}
\end{equation}

\begin{equation}
	\mathbf{M}_{V-V}=softmax( \frac{\mathbf{H}_{p}^V({\mathbf{H}_{t}^V}){}^T}{\sqrt{d}})\mathbf{H}_{t}^V
	\label{eq:commutative9}
\end{equation}

\begin{equation}
	\mathbf{M}_{S-S}=softmax( \frac{({\mathbf{H}_{p}^W\oplus{\mathbf{H}_{p}^V}})({\mathbf{H}_{t}^W}\oplus{\mathbf{H}_{t}^V}){}^T}{\sqrt{d}})(\mathbf{H}_{t}^W\oplus{\mathbf{H}_{t}^V})
	\label{eq:commutative10}
\end{equation}

Where $\mathbf{H}_{p}^W$ and $\mathbf{H}_{t}^W$ represent the text representations of the post and the topic in the latent space, respectively. $\mathbf{H}_{p}^V$ and $\mathbf{H}_{t}^V$ represent the image representations of the post and the topic in the latent space, respectively. $\mathbf{M}_{W-W}$ denotes the semantic relationship feature between the topic text and the post text; $\mathbf{M}_{W-V}$denotes the semantic relationship feature between the topic text and the post image; $\mathbf{M}_{V-W}$denotes the semantic relationship feature between the topic image and the post text; $\mathbf{M}_{V-V}$denotes the semantic relationship feature between the topic image and the post image. These are local semantic relations between topics and posts. $\mathbf{M}_{S-S}$denotes the overall semantic relational feature between topic and post. $\oplus{}$ denotes the concatenation of vectors.

\subsubsection{The Global Fusion Mechanism Based on Discussion Topic}

Typically, noise or misleading information arising during the process of multimodal deep fusion can hinder the extraction of the maximum relevant semantic information, thereby affecting the performance of the model. In order to ensure that the maximum information related to the topic can be extracted and prevent the interference of noise, a topic-based global fusion mechanism is designed. Specifically, local semantic relationships are derived by combining the local features of posts with the local features of topics, while global semantic relationships are established by integrating the overall features of posts with the overall features of topics. Subsequently, five gating mechanisms are employed to ensure that the maximum semantic relations between the post and the topic are output in both the local and global ranges. Finally, the maximum relevant semantic information based on the topic is obtained through the global fusion mechanism based on the topic. 


The five gates are the topic-post text gate $\mathbf{g}_{W-W}$, the topic-post text visual gate $\mathbf{g}_{W-V}$, the topic-post visual text gate $\mathbf{g}_{V-V}$, the topic-post visual gate, and the topic-post overall gate $\mathbf{g}_{S-S}$. Their specific contents are as follows:
\begin{equation}
	\mathbf{g}_{W-W}=[\mathbf{M}_{W-W}\oplus{\mathbf{H}_{t}^W}]\mathbf{W}_{W-W}+\mathbf{b}_{W-W}
	\label{eq:commutative11}
\end{equation}
\begin{equation}
	\mathbf{g}_{W-V}=[\mathbf{M}_{W-V}\oplus{\mathbf{H}_{t}^W}]\mathbf{W}_{W-V}+\mathbf{b}_{W-V}
	\label{eq:commutative12}
\end{equation}

\begin{equation}
	\mathbf{g}_{V-W}=[\mathbf{M}_{V-W}\oplus{\mathbf{H}_{t}^V}]\mathbf{W}_{V-W}+\mathbf{b}_{V-W}
	\label{eq:commutative13}
\end{equation}

\begin{equation}
	\mathbf{g}_{V-V}=[\mathbf{M}_{V-V}\oplus{\mathbf{H}_{t}^V}]\mathbf{W}_{V-W}+\mathbf{b}_{V-V}
	\label{eq:commutative14}
\end{equation}

\begin{equation}
	\mathbf{g}_{S-S}=[\mathbf{M}_{S-S}\oplus{\mathbf{H}_{t}^V}]\mathbf{W}_{S-S}+\mathbf{b}_{S-S}
	\label{eq:commutative15}
\end{equation}
Where $\mathbf{M}_{W-W}$,$\mathbf{M}_{W-V}$,$\mathbf{M}_{V-W}$, $\mathbf{M}_{V-V}$,$\mathbf{M}_{S-S}$ are trainable parameters. Finally, a topic-based representation of maximally relevant semantic information is obtained through a fusion mechanism:
\begin{equation}
    \hat{\mathbf{M}}= \mathbf{M}_{S-S}+\mathbf{M}_{W-W}\otimes{\mathbf{g}_{W-W}}+\mathbf{M}_{W-V}\otimes{\mathbf{g}_{W-V}}+\mathbf{M}_{V-V}\otimes{\mathbf{g}_{V-V}}+\mathbf{M}_{S-S}\otimes{\mathbf{g}_{S-S}}
	\label{eq:commutative16}
\end{equation}

Where $\otimes$ represents matrix multiplication, this vector $\hat{\mathbf{M}}$ is the representation of a local-global semantic correlation feature.

\subsection{The Multi-level Evidential Relational Reasoning}

Generally, when analyzing whether a post accurately answers the question of the discussion topic, the relationship between the post and the topic is analyzed at different levels. First, the analysis considers whether the post responds to the key points of the topic's question, focusing on the significant information relationship between the post and the topic at the macro level. Subsequently, a detailed examination is conducted to explore the fine-grained relationship between the post's content and the topic's question, specifically delving into the intricate connections at the micro level. Therefore, a multi-level evidence relationship reasoning module is designed to capture the complex and subtle relationship between the post and the topic at both macro and micro levels. The first level is the topic-post significant information evidence reasoning, which discovers the connection of key information between the post and the topic at the macro level. The second level is the topic-post internal logic relationship evidence reasoning, which involves a detailed analysis of the internal logical relationships between the topic and the post at the micro level, that is, an in-depth exploration of the relationships between fine-grained contents.

Next, Figure 1 is used as an example to elaborate on the multi-level evidential relational reasoning. Initially, through topic-post significant information evidence reasoning, the most prominent informational connection between the post and the topic is analyzed at the macro level, specifically assessing whether the post addresses the key question of the topic: "Prove whether $A^n$ is an upper triangular matrix using mathematical induction." The analysis reveals that both Post 1 and Post 2 respond to the topic's question. Then, utilizing the topic-post internal logic relationship evidence reasoning, the fine-grained informational relationship between the post content and the topic is examined at the micro level. This includes evaluating the logical coherence of the proof steps, the completeness of the exposition, and the clarity of the language. A detailed analysis of the two posts indicates that the proof process in Post 2 demonstrates clear and well-structured logical relationships, accompanied by comprehensive language expression. Consequently, Post 2 is deemed to provide a superior response and receives a higher score. Thus, the multi-level evidential relational reasoning module, through its analysis at both macro and micro levels, effectively captures the complex and subtle relationships between the post and the topic, facilitating a more nuanced understanding of their interplay.

\subsubsection{The Topic-Post Significant Information Evidence Reasoning}

A topic-post relationship evidence graph is constructed to get the relationship between topic and post, which describes the relationship between the vertices of a finite set by taking each row of features of $\mathbf{H}_{p}^W ,\mathbf{H}_{p}^V ,\mathbf{H}_{t}^W ,\mathbf{H}_{t}^V $ with nodes. To reduce computation and memory consumption, a MLP (Multi-Layer Perceptron) is used to connect each node. In the topic-post significant information relationship evidence graph, a hidden node at the $z-th$ layer can be denoted as $G_{t-p}=\{g_{t-p,1},...,g_{t-p,n}\}$, and the weights of the edges between pairs of nodes are neighbor matrices, which are automatically learned during training. For the topic-post significant information relationship evidence graph, the $i-th$ semantic node at the first layer is initialized as $g_{i}^{0}=[\mathbf{H}_{p,i}^{W},\mathbf{H}_{p,i}^{V},\mathbf{H}_{t,i}^{W},\mathbf{H}_{t,i}^{V}],i\in{\{1,......,2(l_D+l_I)\}}$. For the $z-th$ layer the process of calculating the adjacency matrix between pairs of nodes:
\begin{equation}
    \widetilde{\mathbf{A}}_{i,j}^Z=MLP^{Z-1}([g_{t-p,i}^{Z-1},g_{p,j}^{Z-1}]),i\ne{j}
	\label{eq:commutative17}
\end{equation}

\begin{equation}
    \widetilde{\mathbf{A}}^Z=\sum_{i\in{2(l_D+l_I)} }\sum_{j\in{\mathcal{N}_i} }\widetilde{\mathbf{A}}_{i,j}^Z
	\label{eq:commutative18}
\end{equation}
\begin{equation}
	{\mathbf{A}}^Z=softmax(\widetilde{\mathbf{A}}^Z),  
	\label{eq:commutative19}
\end{equation}

In this process, the adjacency matrix at $z$-th layer is passed from $z-1$-th layer. Where $MLP^{Z-1}$ represents an MLP at the $z-1$-th layer. $\widetilde{\mathbf{A}}_{i,j}^Z$ denotes the semantic coefficient between node $i$ and its neighbour node $j\in{\mathcal{N}_i}$. $\widetilde{\mathbf{A}}^Z$ is normalised by a softmax function to get ${\mathbf{A}}^Z$. Next, the semantic nodes are computed at $z-th$ level:
\begin{equation}
	\mathbf{g}_{t-p,j}^{Z}=\sum_{j\in{\mathcal{N}_i} }\mathbf{A}_i^{Z} \mathbf{g}_{t-p,j}^{Z-1}
	\label{eq:commutative20}
\end{equation}

Next, the topic-post relationship graph by stacking L inference layers is obtained for message passing between different semantic nodes. The hidden state of the topic-post relationship graph for the L-th layer is represented as $\mathbf{g}_{t-p,j}^{L}$. Then, the final hidden states of the topic-post relationship graph are aggregated to obtain the topic-post relationship graph embedding.
\begin{equation}
   \mathbf{G}_{t-p}=Sum(\mathbf{g}_{t-p,*}^{L})
	\label{eq:commutative21}
\end{equation}

Subsequently, the retrieval-based attention mechanism is employed to capture the relationship graph-oriented attention information from the concatenation of topic and post representations $\mathbf{H}_{t-p}=[\mathbf{H}_t^{W},\mathbf{H}_t^{V},\mathbf{H}_p^{W},\mathbf{H}_p^{V}]=\{h_1,h_2,...,h_{2(l_D+l_I)}\}$ and the topic-post relationship graph representation $G_{t-p}$. It aims to retrieve obvious links between the topic-post relationship graph and the topic-post representations. Attention weights are calculated as:
\begin{equation}
	\alpha_j=\frac{exp(\beta_j)}{ \sum_{S=1}^{2(l_D+l_I)}exp(\beta_S)} )
	\label{eq:commutative22}
\end{equation}
\begin{equation}
	\beta_j=\sum_{i\in \mathcal{C}}h_{t-p,j}^{\top}G_{t-p,i}
	\label{eq:commutative22_2}
\end{equation}
Where $\mathcal{C}$ denotes the index set of feature nodes in the post-topic relationship graph. ${\top}$ represents the matrix transposition. The final topic-post significant information relationship feature is denoted as:
\begin{equation}
	G_{SIG}=\sum_{}\alpha_{t-p}\mathbf{H}_{t-p}
	\label{eq:commutative23}
\end{equation}

\subsubsection{The Topic-Post Internal Logic Relationship Evidence Reasoning}

To analyze the complex relationship between posts and discussion topics at micro level, a topic evidence graph is first constructed utilizing the entire discussion topic. This graph helps in exploring the topic's internal relationship. Then, the final features from the topic evidence graph are combined with the post features to construct a topic-post internal relationship evidence graph, which delves into the subtle and complex relationship between the topic and within the post. Therefore, a post-evidence graph is initially constructed, describing the relationship between the vertices of a finite set by taking each row of features of $\mathbf{H}_{p}^W$,$\mathbf{H}_{p}^V$ with $l_D+l_I $ nodes. The two nodes are connected using MLP. The topic evidence graph is constructed in the same way. The $z$-th level post evidence graph and discussion topic evidence graph can be represented as  $G_{t}^Z=\{g_{t,1}^Z,...,g_{t,n}^Z\}$ and $G_{p}^Z=\{g_{p,1}^Z,...,g_{p,n}^Z\}$, respectively. The weights of edges connected between semantic nodes using MLP are adjacency matrices, which are automatically learned during training. Next, the process of building the post-evidence graph $G_p$ is illustrated as an example. For the first layer the $i$-th semantic node initialised as $g_{i}^{0}=[\mathbf{H}_{p,i}^{W},\mathbf{H}_{p,i}^{V}],i\in{\{1,......,l_D+l_I\}}$. Then, the process of computing the adjacency matrix $A^Z$ for the $z$-th layer:

\begin{equation}
    \hat{\mathbf{A}}_{i,j}^Z=MLP^{Z-1}([g_{p,i}^{Z-1},g_{p,j}^{Z-1}]),i\ne{j}
	\label{eq:commutative24}
\end{equation}

\begin{equation}
	{\mathbf{A}}_{p}^Z=softmax(\hat{\mathbf{A}}^Z),  
	\label{eq:commutative25}
\end{equation}
Where $\hat{\mathbf{A}}_{i,j}^Z$denotes the semantic coefficient between node i and it's neighbour node j. $\hat{\mathbf{A}}^Z$ is normalized by a softmax function to ${\mathbf{A}}_{p}^Z$. Subsequently, the semantic nodes in $z$-th layer are computed as follows:
\begin{equation}
	\mathbf{g}_{p,j}^{Z}=\sum_{j\in{\mathcal{N}_i} }\mathbf{A}_i^{Z} \mathbf{g}_{p,j}^{Z-1}
	\label{eq:commutative26}
\end{equation}

The internal relations of a discussion topic or a post is obtained by stacking L inference layers. The hidden states of the $L$-th layer inference graphs of discussion topics and posts are denoted as $\mathbf{g}_{p,n}^{L}$  and $\mathbf{g}_{t,n}^{L}$ respectively. The last hidden state of the evidence graph of a discussion topic is aggregated, i.e., the correlation graph embedding of the discussion topic. It is used as a feature to learn its relationship with posts.
\begin{equation}
   \mathbf{t}=Aggregate(\mathbf{g}_{t,*}^{L})
	\label{eq:commutative27}
\end{equation}

\begin{equation}
 {R}_i=softmax(MLP([\mathbf{t},\mathbf{g}^{L}_{p,i}]))
	\label{eq:commutative28}
\end{equation}

Where $Aggregate$ is an aggregation operation to derive the discussion topic evidence graph embedding $\mathbf{t}$. $MLP$ is used to compute the coherence relationship weights ${R}_i$ between the discussion topic coherence graph embedding and the $i$-th semantic node within the post. Subsequently, the correlation relationship weights are multiplied with the final post-evidence graph to ultimately obtain the topic-post relationship feature $G_{TPL}$:

\begin{equation}
   \mathbf{G}_{TPL}=\sum_{i}R_{i}\mathbf{g}^{L}_{p,i}
	\label{eq:commutative30}
\end{equation}
Finally, two levels of evidence features are concatenated:
\begin{equation}
	\mathbf{R}_{MEG}=\mathbf{G}_{SIG}\oplus{\mathbf{G}_{TPL}}
	\label{eq:commutative31}
\end{equation}

\subsection{Post Quality Score Prediction}


To obtain the final post-quality correlation feature, the local-global semantic correlation reasoning feature $\hat{\mathbf{M}}$ is combined with the multi-level evidence graph reasoning feature $R_{MEG}$. Subsequently, this post-quality correlation feature is passed to a linear layer to derive the ultimate post-quality score:

\begin{equation}
    F(T_{i},p_{i,j})={W}_{p}([\hat{\mathbf{M}}, \mathbf{R}_{MEG}])+{b}_p
	\label{eq:commutative32}
\end{equation}
Where $\mathbf{W}_{p}$ is the projection parameter. ${b}_p$ is the bias term. $T_i$ signifies the $i$-th discussion topic, and $p_{i,j}$ corresponds to the $j$-th post beneath the $i$-th discussion topic, denoted as $T_i$. The loss function used is given by:
\begin{equation}
	L=\sum_{i}\max(0,\gamma-F(T_{i},p^{+})+F(T_{i},p^{-}))
	\label{eq:commutative33}
\end{equation}
Where $p^{+},p^{+}\in{P_i}$ is the post quality score in a given discussion topic, i.e., $p^{+}$ has a higher relevance score than $p^{-}$. $\gamma$ is a scale factor that magnifies the difference between the fraction and the margin.

\section{Experimental Setup}
\subsection{Datasets}

Existing research on post-quality assessment primarily emphasizes textual data, revealing a notable absence of a multimodal dataset for post-topic. In the field of education, achieving accurate and comprehensive assessment has consistently been a primary objective for educators, underscoring the necessity of utilizing multimodal data to thoroughly evaluate post quality. The creation of a multimodal topic-post dataset not only improves the precision of post-quality assessment but also offers significant potential for application in related educational research areas, such as the evaluation of short-answer questions. MOOC is an online learning platform with a large number of learners and supports multiple forms of answering questions, resulting in a vast amount of multimodal data. The topic-post pair data was crawled from the MOOC platform of Chinese universities, and three multimodal datasets were constructed. The process of dataset construction is as follows: (1) Crawling multimodal topic-post pairs. The initial step involves using course categories as the main topic and traversing all courses within specific categories. For each course, all associated discussion topics are iterated through, and topic-post pairs within each discussion topic are captured. This approach establishes the correspondence among course categories, courses, topics, and posts. To ensure the quality of the multimodal dataset, an automatic analysis is conducted during the crawling process to determine whether the topic-post pairs constitute multimodal data. Pairs that consist solely of plain text data are excluded. (2) Data preprocessing. Initially, topic-post pairs with empty content are filtered out. Subsequently, the dataset is scrutinized for duplicate topic-post pairs. Finally, special characters and emojis are removed from the text. (3)Expanding dataset size through text rewriting and image enhancement Techniques. For text data, sentence-level text rewriting is employed to alter the representation of the text without changing its original meaning. Specifically, the text of the discussion topic and post is initially translated into one language, followed by a subsequent translation into a second language. This process involves two consecutive translations into different languages, and ultimately results in the text being translated back into the original language, yielding a final sentence-level text rewriting. Regarding image augmentation, we implement all methods from Rand Augment, with the exception of color inversion and cutout. Color inversion can disrupt the correspondence between the image and the text, while cutout may eliminate small but significant details from the image. (4) Data annotation. Initially, the quality of the posts was classified into 5 levels, $S_{i,j}=\{0,...S\}$. Higher scores indicated more comprehensive answers to the topic's questions. Subsequently, the data was pre-annotated using a multimodal correlation analysis model\cite{fan2019product}, followed by manual proofreading for five individuals. This process resulted in the creation of three datasets: the Art History Course dataset, the Education Course dataset, and the Excellence Course dataset. These datasets are illustrated as follows.


\textbf{1.The Art History Course Dataset}


This dataset comprises courses that pertain to art and history, organized into 14 distinct categories: creative design, art, theater and film, design, music, dance, literature and culture, journalism and communication, philosophy, history, politics, law, civics, and social studies. It includes a total of 8,896 topic-post pairs and 548 discussion topics.

\textbf{2.The Education Course Dataset}


The dataset encompasses data from eight distinct categories: education, teaching ability, information technology instruction, professionalism, subject-specific teaching, quality education, physical education, and preschool education. It includes a total of 24,142 topic-post pairs and 766 discussion topics.

\textbf{3.The Excellence Courses Dataset}


This dataset comprises a curated selection of courses across various fields, including science, agronomy, computer science, economics, and management. It consists of 24,775 posts organized into 1,485 topics.  Table 4 presents detailed statistics for the three datasets involved. Specifically, the test set comprises 20$\%$ of the entire dataset, while the training set makes up 80$\%$. Additionally, the validation set is designated as 20$\%$ of the training set.
\begin{table}[htbp]
	\centering
	\begin{tabular}{  c | c  c  }
		
		\hline
		\toprule[2pt]
		\textbf{Datasets} &  \multicolumn{2}{c}{\textbf{Instance Number (\#{T}/\#{P})}}     \\ 
		\textbf{} &  \textbf{Train+Dev} &\textbf{Test}   \\ \hline
	    The Education Course&	526/19142&	240/5000 \\
	    The Excellence Courses&	1057/18775&	428/5000 \\
		The Art History Course&	266/6428&	282/2468 \\
		\bottomrule[2pt]
	\end{tabular}
	\caption{Statistics of the three datasets where \#{T} and \#{P} represent the number of discussion topics and posts, respectively.}
\end{table}
\begin{figure}[htbp]
	\centering
	\begin{minipage}{0.65\linewidth}
		\centering
		\includegraphics[width=1\linewidth]{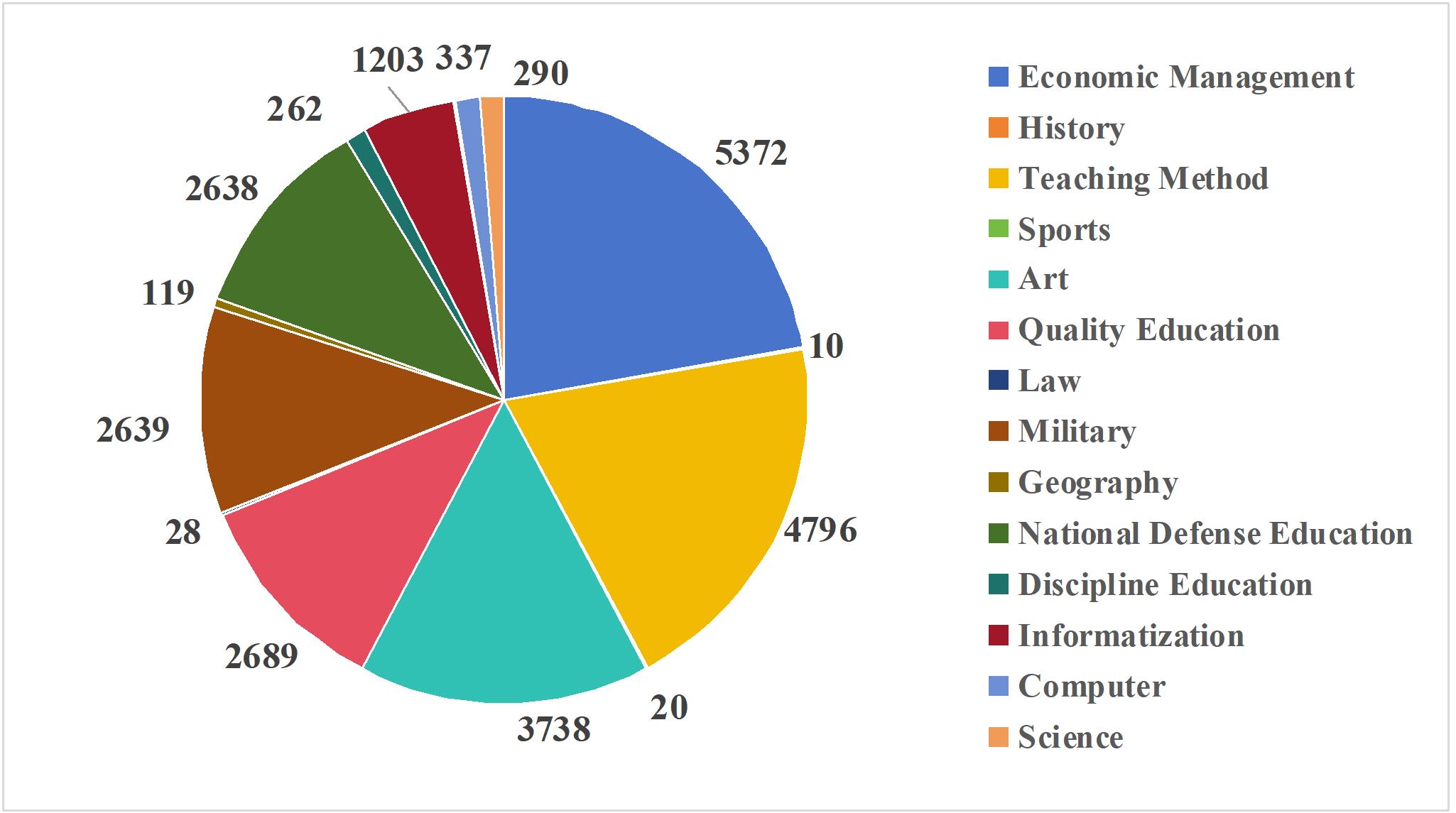}
		\caption{The Excellent Course dataset}
		\label{chutian12}
	\end{minipage}
	\begin{minipage}{0.65\linewidth}
		\centering
		\includegraphics[width=1\linewidth]{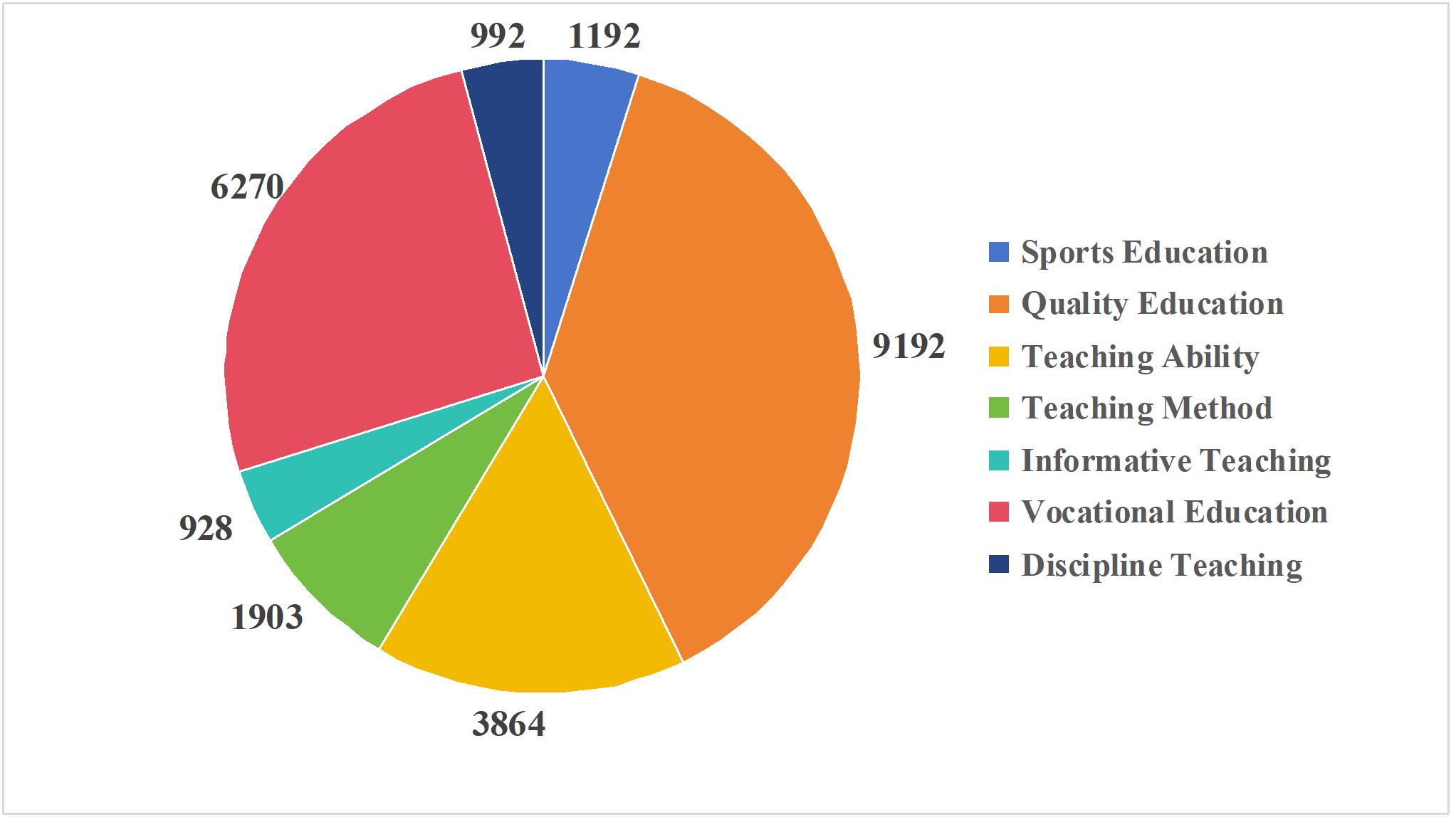}
		\caption{The Education Course dataset}
		\label{chutian23}
	\end{minipage}
   \begin{minipage}{0.65\linewidth}
   	\centering
   	\includegraphics[width=1\linewidth]{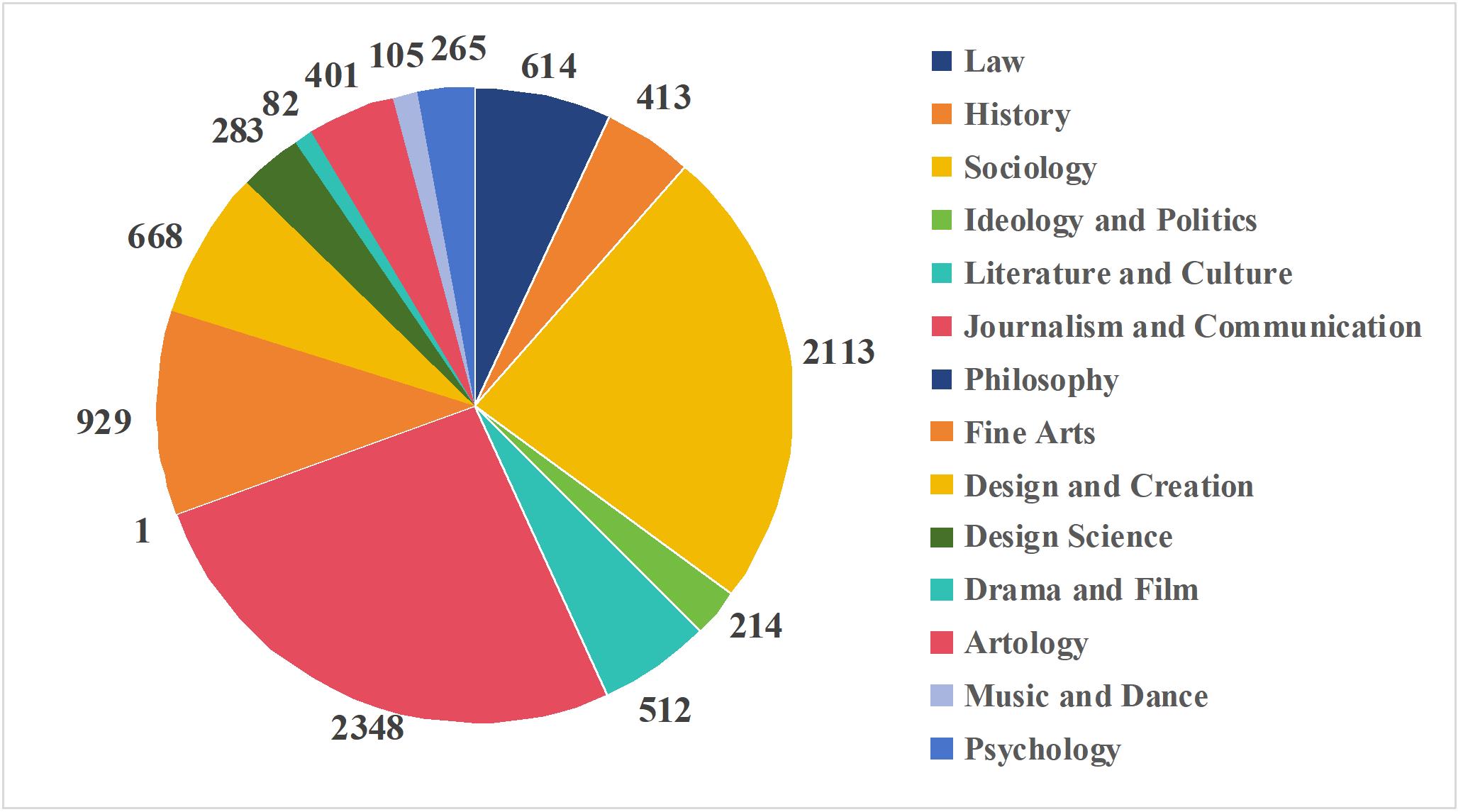}
   	\caption{The Art History Course dataset}
   	\label{chutian24}
   \end{minipage}
\end{figure}

\textbf{Quantitative Analysis}

An analysis was conducted to examine the diversity of data categories across the three datasets. These datasets comprise 56,813 multimodal topic-post pairs, encompassing 9 major categories and 110 subcategories, and spanning fields such as science, engineering, agriculture, foreign languages, economics and management, computers, music, and art. To provide a clearer representation of the percentage of different categories in each dataset, pie charts have been utilized for visualization, as depicted in Figures 5, 6, and 7. These graphs notably reveal that the Education category has the highest count of topic-post pairs.

\textbf{Quality Analysis}

The quality of the datasets is directly influenced by the proportion of multimodal data they contain. Upon analysis, it was discovered that all three datasets comprised exclusively of 100$\%$ multimodal data. This high percentage is a result of the data crawling process, which retained only topic-post pairs that included multimodal data types, thereby excluding unimodal data.

\subsection{Implementation Details}

To ensure fair comparison methods, the same processing method is applied to the data. For image data, the pre-trained network CSPDarkNet is used to extract multi-scale visual features. In the case of text data, the ICU tokenizer is first employed to segment the text, followed by the initialization of word vectors using fastText. In the network configuration, the n-gram kernel sizes for text are set to 1, 3, and 5. The dimension of the text hidden layer is configured to 128, and the encoding size of the image representation's feature, denoted as $d_{l_D}$, is also established at 128. Within the multilevel evidence graph inference module, the number of graph inference layers is configured to 2, with each hidden layer dimension set to 128. During the training process, the batch size is set to 4, and the optimizer uses Adam's optimizer. The hyperparameter $\gamma$ in the loss function is set to 1.

\subsection{Compared Methods}

The model is compared to six state-of-the-art correlation analytical methods using textual and multimodal data. The comparison methods using only textual data include the Bilateral Multi-Perspective Matching (BiMPM) model \cite{wang2017bilateral}, the Convolutional Kernel Based Neural Ranking Model (Conv-KNRM) \cite{dai2018convolutional}, and the Cross-Domain Analysis of Posts model (CDAP) \cite{capuano2021transfer}. A review of the related literature reveals a lack of studies employing multimodal data to assess the quality of posts. Therefore, the models are compared with three strong inference models that utilize multimodal data: Stochastic Shared Embeddings Cross-attention (SSE-Cross) \cite{abavisani2020multimodal}, which uses stochastic shared embeddings and cross-modal attention mechanisms to merge different modalities for reasoning; Multi-perspective Coherent Reasoning (MCR)\cite{liu2021multi}, which uses multimodal data for intra- and inter-modal analysis; and Dual Incongruity Perceiving (DIP) \cite{wen2023dip}, which extracts multimodal data from factual and affective levels to identify valuable information for reasoning; $G^2SAM$ \cite{wei2024g} constructs multimodal graphs for each instance and maps them into a semantic space. It then builds a relationship graph among different instances and performs reasoning based on semantic correlations.

\subsection{Evaluation Metrics}

In this paper, the output of MFTRR is a list of posts ranked by their quality scores. Two commonly used ranking-based metrics are adopted to assess the performance of the mode: Mean Average Precision (MAP) and Normalized Discounted Cumulative Gain (NDCG@N). MAP is a widely recognized evaluation metric that assesses the overall ranking performance of a set of candidate posts. NDCG@N evaluates the ranking performance based on the quality scores of the top N posts, particularly in contexts where teachers typically review only a limited number of posts. Generally, teachers examine the quality of the top N posts to gauge students' understanding of the knowledge point. Consequently, N in NDCG@N is set to 3 and 5.

\section{Experimental Results }
\subsection{Main Results}
\begin{table}[htbp]
	\centering
	\begin{tabular}{  c | c | c  c  c }
		\toprule[2pt]
		\textbf{Type} &\textbf{Method} &  \multicolumn{3}{c}{\textbf{The Excellence Course}}     \\ 
		 & & \textbf{MAP} &\textbf{N@3} &\textbf{N@5}    \\ \hline
		Text-only&CDAP& \textbf{96.28}&51.50&	54.45\\
		  & BiMPM&79.78&79.79&79.88 \\
		  & Conv-KNRM&79.90&79.77&79.77 \\ \hline
		 {Multimodal} & SSE-Cross&85.00&80.10&81.19 \\
		  & DIP&85.05&80.08&82.10 \\
          & MCR&85.11&81.12&82.52 \\
          & $G^2SAM$ & 87.79&	81.10&	82.57 \\
		  & The current method&86.56&\textbf{85.07}&\textbf{85.15} \\
		\bottomrule[2pt]
	\end{tabular}
	\caption{Post quality score prediction results on the Excellence Course dataset.}
\end{table}

\begin{table}[ht]
	\centering
	\begin{tabular}{  c | c | c  c  c }
		\toprule[2pt]
		\textbf{Type} &\textbf{Method} &  \multicolumn{3}{c}{\textbf{The Education Course}}     \\ 
		& & \textbf{MAP} &\textbf{N@3} &\textbf{N@5}    \\ \hline
		Text-only&CDAP&	\textbf{96.51}&49.94&45.53\\
		& BiMPM&79.88&79.79&79.79 \\
		& Conv-KNRM&79.91&79.76&79.76 \\ \hline
		{Multimodal} & SSE-Cross&82.03&84.00&85.00 \\
		& DIP& 87.51&84.62&	86.12 \\
        & MCR& 87.90&85.40&	86.80 \\
        & $G^2SAM$ & 88.51&	85.03&	86.88 \\
		& The current method&89.06&	\textbf{87.10}&	\textbf{88.98} \\
		\bottomrule[2pt]
	\end{tabular}
	\caption{Post quality score prediction results on the Education Course dataset.}
\end{table}

\begin{table}[htbp]
	\centering
        \small 
	\begin{tabular}{  c | c | c  c  c }
		\toprule[2pt]
		\textbf{Type} &\textbf{Method} & \multicolumn{3}{c}{\textbf{The Art History Course}}     \\ 
		& & \textbf{MAP} &\textbf{N@3} &\textbf{N@5}    \\ \hline
		Text-only&CDAP&	\textbf{97.91}&	62.45&72.86\\
		& BiMPM&88.98&82.76&84.45 \\
		& Conv-KNRM&89.44&84.10&85.87 \\ \hline
		{Multimodal} & SSE-Cross&95.18&	84.37&88.82 \\
		& DIP& 95.65&83.26&91.67 \\
        & MCR& 95.88&91.77&	93.14 \\
        & $G^2SAM$ & 96.03&	91.69&	93.20 \\
		& The current method&96.96&\textbf{93.62}&\textbf{94.26} \\
		\bottomrule[2pt]
	\end{tabular}
	\caption{Post quality score prediction results on the Art History course dataset.}
\end{table}

Tables 5, 6, and 7 show the results of the MFTRR model and comparison methods on three datasets. It is observed that the MFTRR model outperforms the other methods on three datasets. An examination of the data within these tables reveals similar trends across the methods. Specifically, methods using multimodal data reasoning significantly outperform those using single-modal data, indicating that adding features can provide more complementary information. Among the comparison methods using textual data, the CDAP method performs well on the MAP metric but weakly on the NDCG@3 and NDCG@5 metrics. This disparity may stem from several factors: (1) The MAP metric is not sensitive to ranking. (2) The CDAP model, which is a transfer learning model trained on education-related data, is better suited for predicting educational posts. Analysis of dataset diversity reveals a high proportion of education-related data in all three datasets. (3) overall, CDAP's performance is the lowest among the comparison methods that exclusively utilize textual data, likely because it focuses solely on analyzing posts during reasoning and analysis without considering the relationships between posts and topics. Additionally, the use of transfer learning in CDAP to predict post-quality scores may be constrained to specific domains due to the influence of the pre-trained model. In the experimental comparison of BiMPM and Conv-KNRM, it was observed that their results on the larger datasets, namely the Education Course and the Excellence Course, were remarkably similar. This finding suggests that both models effectively leverage textual information to capture the relationships between posts and topics when sufficient data is available. However, on relatively smaller datasets, Conv-KNRM outperformed BiMPM. This may be attributed to the strong nonlinear fitting capability of the Gaussian kernel function in Conv-KNRM, which enables it to better analyze similarities even with limited data. Among multimodal comparison methods, our approach outperforms other methods using multimodal data. The primary reasons for this superiority include: (1) Our method simulates the human thinking process from multiple scales and levels to assess the quality of posts; (2) During modal fusion, it analyzes the semantic relationships between posts and topics across different scales while effectively filtering out noise; (3) In analyzing the intricate relationships between posts and topics, it thoroughly explores subtle relationships at different levels. Additionally, several factors affect the effectiveness of the DIP method: (1) DIP primarily evaluates the quality of posts through facts and emotions. however, most posts exhibit a neutral emotional state. (2) DIP was originally designed for binary classification tasks, which limits the model's ability to effectively predict categories with complexity or subtle differences. (3) To facilitate fair comparisons among different models, we have modified DIP into a ranking task, which also exerts a certain influence on the model. SSE-Cross exhibits the lowest performance among the multimodal comparison models for the following reasons: (1) It neglects to consider the relationship between posts and topics, focusing solely on extracting useful information from the text and images within the post. (2) Its analysis is limited to a singular perspective of the fusion of text and images, while the DIP method analyzes the relationship between post images and text from both factual and emotional perspectives. $G^2SAM$ exhibits weaker performance compared to our model, likely because it relies solely on the multimodal data of posts, thereby analyzing the relationships between posts without adequately addressing the critical relationship between topics and posts. The data presented in the table indicates that $G^2SAM$'s performance is relatively close to that of MCR. This similarity may be attributed to two factors: (1) the MAP metric may not be sufficiently sensitive to differences in ranking; (2) $G^2SAM$ enhances semantic relevance by incorporating label-aware graph contrastive learning into the representation of the semantic space.

\begin{table}[b!]
	\centering
	\begin{tabular}{ c | c | c  c  c }
		
		\hline
		\toprule[2pt]
		\textbf{Datasets} &\textbf{Text Feature Extractor} & \textbf{MAP}  & \textbf{N@3} & \textbf{N@5}    \\  \hline
	
		The Education Course&BERT-Base-Chinese&88.21&82.19&84.51\\
        {}&Chinese-BERT-wwm-ext&87.88&82.78&84.58 \\ 
        {}&Chinese-roberta-wwm-ext-large&87.98&82.52&84.28 \\ 
		{}&CNN&\textbf{89.06}&\textbf{87.10}&\textbf{88.98} \\ \hline
		
		{The Excellence Courses}&BERT-Base-Chinese&86.39&83.29&84.11 \\
        {}&Chinese-BERT-wwm-ext&86.29&84.32&84.14 \\ 
        {}&Chinese-roberta-wwm-ext-large&86.20&83.34&84.16 \\ 
		{}&CNN&\textbf{86.56}&\textbf{85.07}&\textbf{85.15} \\ \hline
		{The Art History Course}&BERT-Base-Chinese&96.32&92.65&93.44\\
        {}&Chinese-BERT-wwm-ext&96.30&92.77&93.46 \\ 
         
		{}&CNN&\textbf{96.96}&\textbf{93.62}&\textbf{94.26} \\
		\bottomrule[2pt]
	\end{tabular}
	\caption{Experimental results using different text feature extractors.}
\end{table}

\subsection{Comparison of Text Feature Extraction Methods}

In this paper, Convolutional Neural Networks (CNNs) are used to learn contextual text representations. C-gram CNNs can learn multiple ranges of feature relationships with different-sized convolutional kernels, while the features extracted by CNNs are both neutral and efficient. In Natural Language Processing (NLP) research, some work uses BERT (BERT-Base-Chinese) \cite{devlin2018bert}, Chinese-BERT-wwm-ext \cite{cui2021pre}, and  Chinese-roberta-wwm-ext-large as feature extractors. A comparison experiment was conducted by replacing the feature extractors with BERT-Base-Chinese, Chinese-BERT-wwm-ext, and Chinese-roberta-wwm-ext-large, and the results of the experiment are shown in Table 8. The data reveals that CNN yields higher results compared to the other three. This could be attributed to CNN's ability to learn a wide range of textual relational features using different convolutional kernels. Furthermore, the features extracted by CNNs, which learn varying ranges of textual relationships, can be more deeply and multilayered fused with multi-scale visual features during cross-modal interactions. When comparing BERT-Base-Chinese and Chinese-BERT-wwm-ext, it is evident that their performances are very close, with Chinese-BERT-wwm-ext exhibiting a slight advantage over the former. The proximity in performance may be attributed to the models' inherent ability to effectively learn the complex relationships between posts and topics, thus minimizing the impact of different feature extraction methods on model efficacy. The slightly better performance of Chinese-BERT-wwm-ext can be ascribed to its foundation in BERT, coupled with more extensive pre-training on a richer corpus and the utilization of more advanced pre-training techniques, enabling it to extract a greater number of generalized features.

\subsection{Analysis of the Structure of Local-global Semantic Correlation Reasoning Module}

\begin{figure}[b!]
	\centering
	\begin{minipage}{0.75\linewidth}
		\centering
		\includegraphics[width=0.97\linewidth]{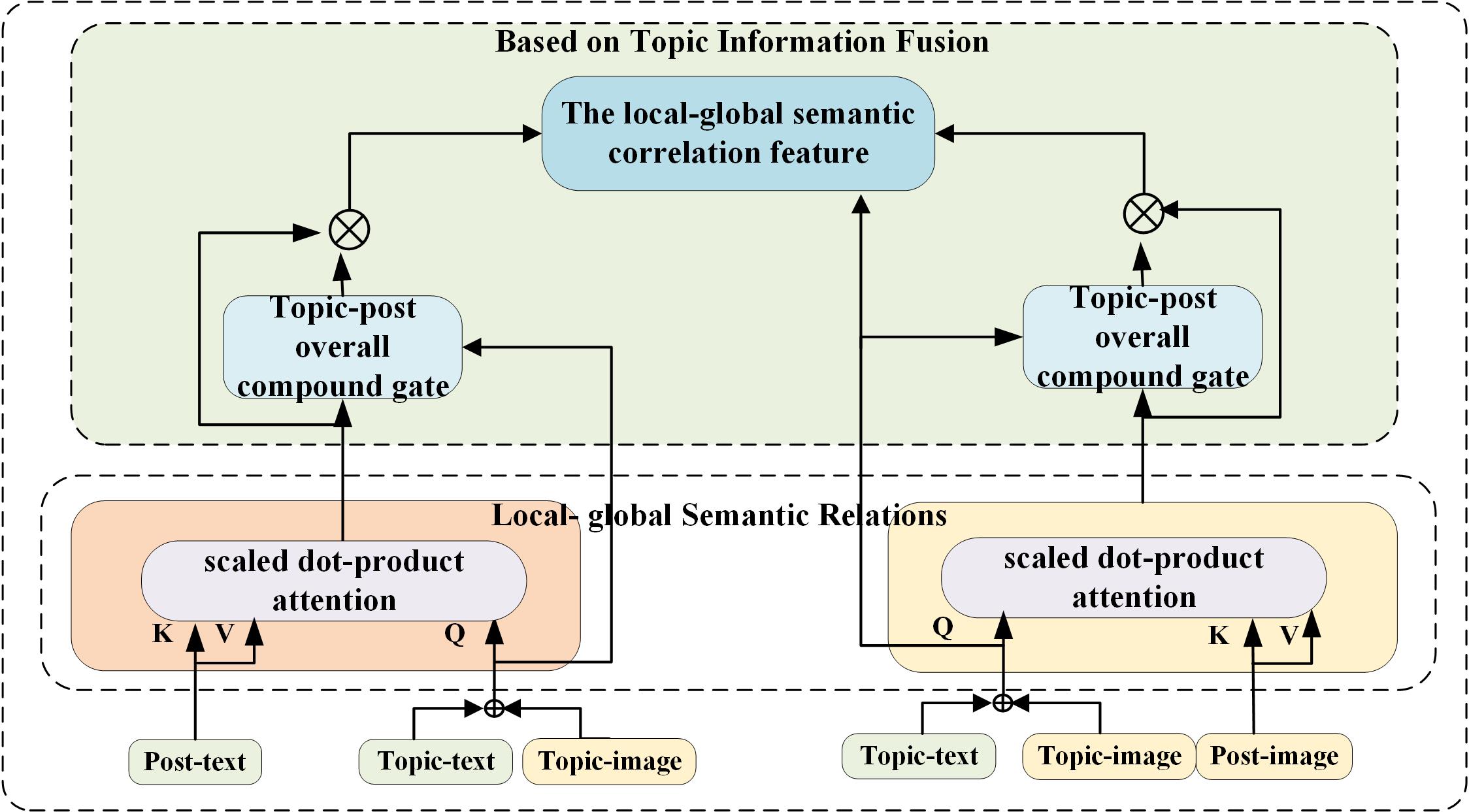}
		\caption{The structure of MFTRR(I)}
		\label{chutian11}
	\end{minipage}
\end{figure}

\begin{figure}[htbp]
	\centering
 \begin{minipage}{0.5\linewidth}
   	\centering
   	\includegraphics[width=0.78\linewidth]{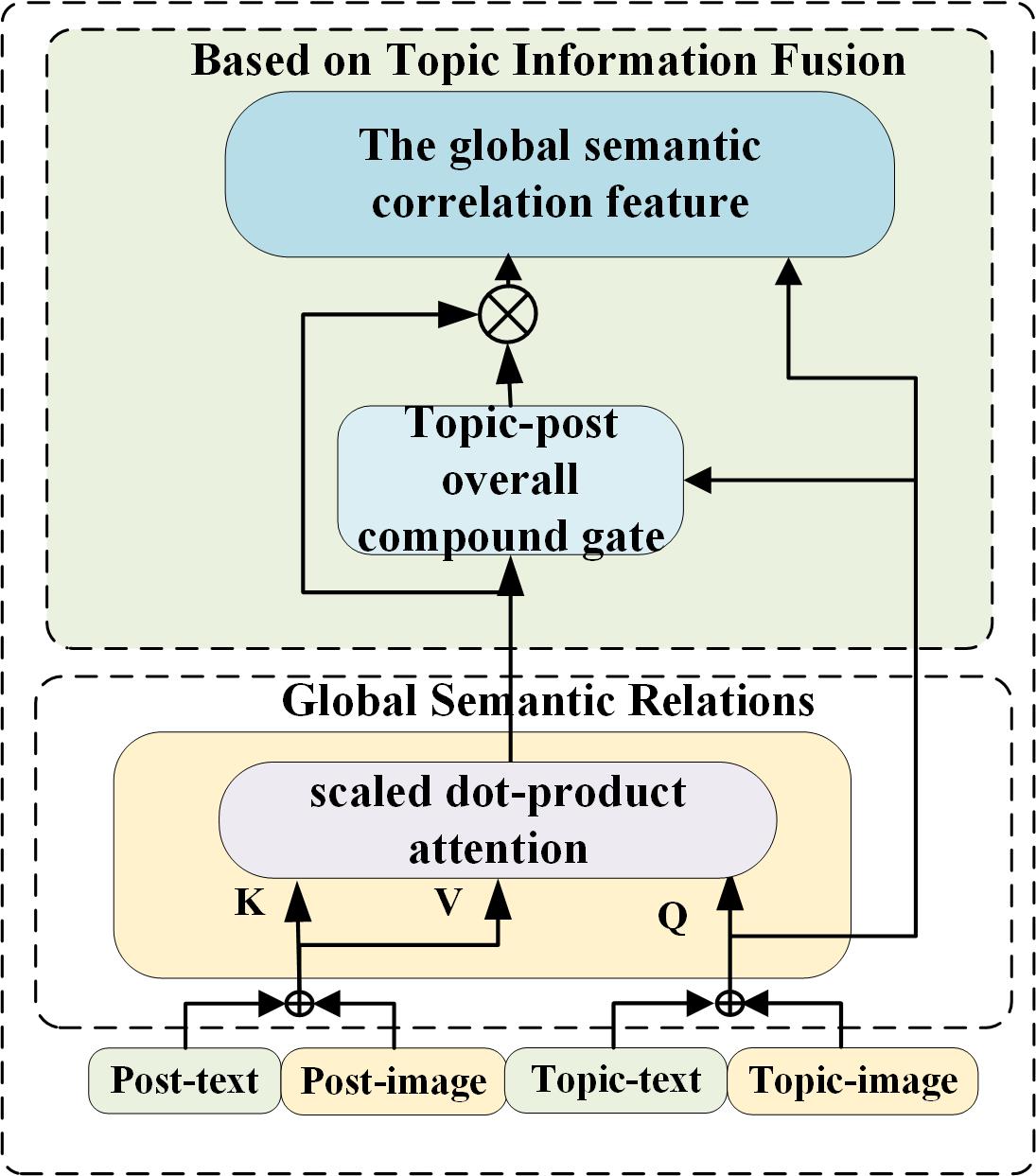}
   	\caption{The structure of MFTRR(II)}
   	\label{idea211}
   \end{minipage}
    \begin{minipage}{0.96\linewidth}
   	\centering
   	\includegraphics[width=1\linewidth]{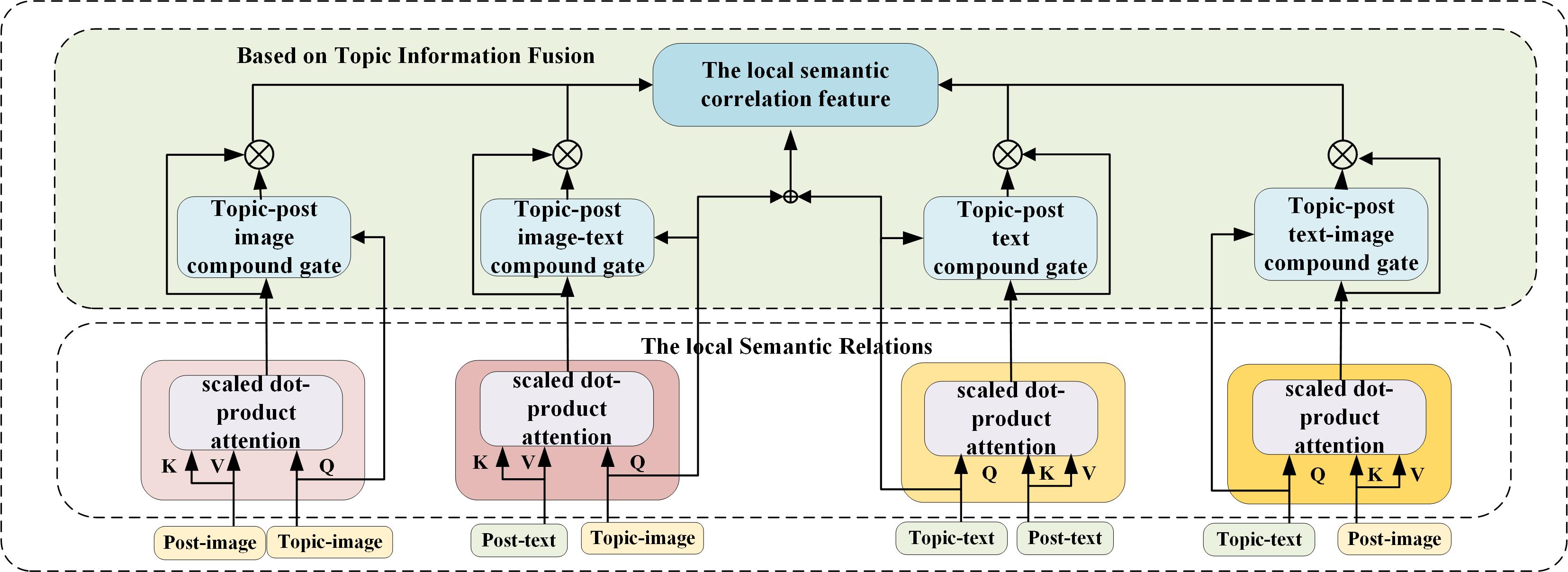}
   	\caption{The structure of MFTRR(III)}
   	\label{idea213}
   \end{minipage}
   \begin{minipage}{0.96\linewidth}
   	\centering
   	\includegraphics[width=1\linewidth]{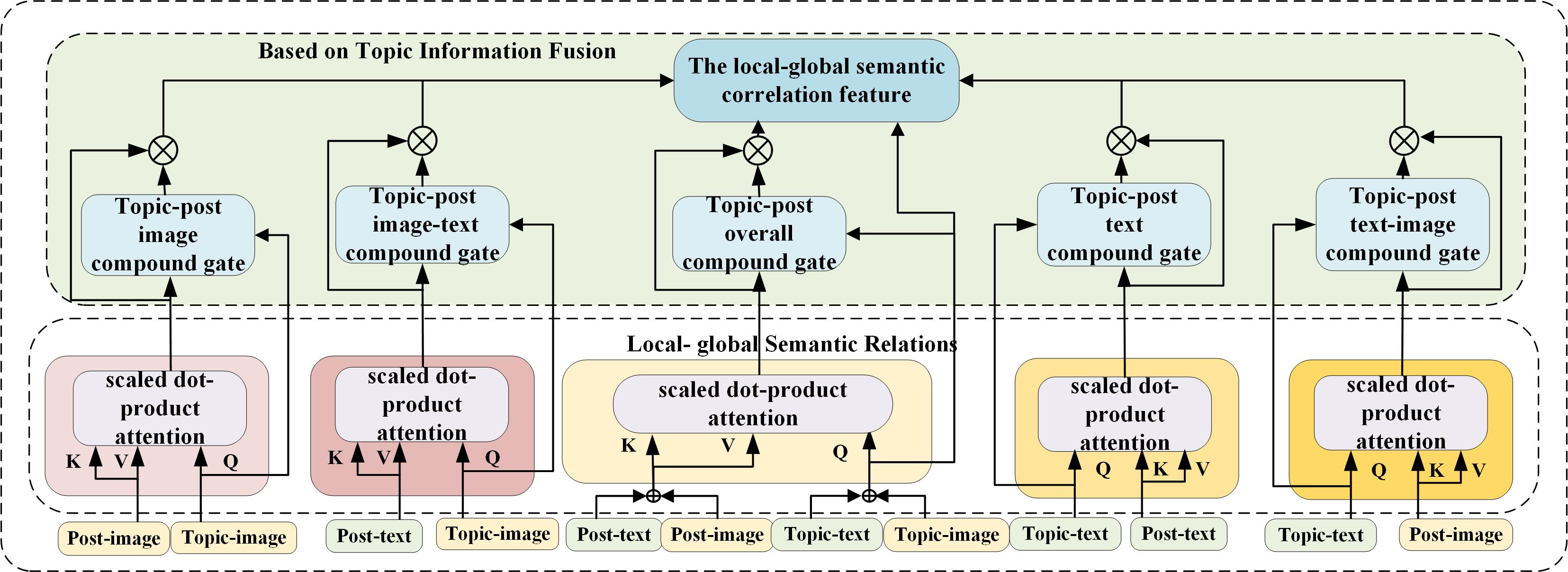}
   	\caption{The structure of MFTRR(IV)}
   	\label{idea214}
   \end{minipage}
\end{figure}
Four distinct structures of the Semantic Correlation Reasoning Module were designed to compare their effects. Structure 1, illustrated in Figure 8, explores the semantic relationship between the different modalities of a post and the overall text- and image-based discussion topic, denoted as MFTRR(I). Structure 2, depicted in Figure 9, is the semantic relationship between the discussion topic and the post based on analyzing text and images as a whole, denoted as MFTRR(II). Structure 3, shown in Figure 10, explores the semantic relationship between topics and posts at the local modal information level. Structure 4, represented in Figure 11, analyzes the semantic relationship between topics and posts in terms of both localized modality and based on overall information, denoted as MFTRR(IV). Comparative experiments were conducted on the three datasets, with the results presented in Table 9. The experimental results indicate that MFTRR(IV) outperforms the other structures. It demonstrates that the structure adequately analyses the semantic relationship between posts and topics from multiple scales.

\begin{table}[htbp]
	\centering
	\begin{tabular}{  c | c | c  c  c }
		
		\hline
		\toprule[2pt]
		\textbf{Datasets} &\textbf{Model Variant} & \textbf{MAP}  & \textbf{N@3} & \textbf{N@5}    \\  \hline
	
		The Education Course&MFTRR(I)&88.07&85.50&86.54\\
		{}&MFTRR(II)&87.96&85.55&86.15 \\ 
        {}&MFTRR(III)&88.73&85.19&86.14 \\ 
		{}&MFTRR(IV)&\textbf{89.06}&\textbf{87.10}&\textbf{88.98} \\ \hline
		{The Excellence Courses}&MFTRR(I)&86.23&84.52&84.94 \\
		{}&MFTRR(II)&86.24&84.54&84.97 \\ 
        {}&MFTRR(III)&86.23&84.32&84.87 \\ 
        {}&MFTRR(IV)&\textbf{86.56}&\textbf{85.07}&\textbf{85.15} \\ \hline
		{The Art History Course}&MFTRR(I)&96.90&93.57&94.02\\
		{}&MFTRR(II)&96.82&93.52&93.89 \\
        {}&MFTRR(III)&96.82&93.51&93.93 \\
        {}&MFTRR(IV)&\textbf{96.96}&\textbf{93.62}&\textbf{94.26} \\
		\bottomrule[2pt]
	\end{tabular}
	\caption{The Local-global Semantic Correlation Reasoning modules that different structures.}
\end{table}

   

\subsection{Analysis of the Structure of Multi-Level Evidence Relational Reasoning Module}

Graph Convolutional Neural Networks (GCNs) have been extensively employed in various research studies to analyze relational information \cite{liang2022multi}. Nonetheless, the training process of GCNs typically demands substantial memory and computational resources, and the relational aspect analyzed by GCNs is relatively simplistic. In this study, the multi-level evidential relational reasoning module was replaced with a two-layer GCN, referred to as MFTRR(GCN). Comparative experiments were conducted, and the results are presented in Table 10. The data clearly indicate that MFTRR significantly outperforms MFTRR(GCN). This could be attributed to the comprehensive analysis of the relationships between posts and discussion topics across multi-levels. When analyzing whether a post adequately answers the question of the topic, the consistency of key information between the post and the discussion topic is first checked, followed by a careful examination of the relationship between the overall content of the post and the topic. This approach facilitated a more nuanced analysis of the post's relationship with the discussion topic across various levels.


\begin{table}[ht]
	\centering
	\begin{tabular}{  c | c | c  c  c }
		
		\hline
		\toprule[2pt]
		\textbf{Datasets} &\textbf{Model Variant} & \textbf{MAP}  & \textbf{N@3} & \textbf{N@5}    \\  \hline
	
		The Education Course&MFTRR&\textbf{89.06}&\textbf{87.10}&\textbf{88.98}\\
		{}&MFTRR(GCN)&{68.41}&{42.11}&{47.18} \\ \hline
		
		{The Excellence Courses}&MFTRR&\textbf{86.56}&\textbf{85.07}&\textbf{85.15}\\
		{}&MFTRR(GCN)&{68.41}&{42.11}&{47.18} \\ \hline
		{The Art History Course}&MFTRR&\textbf{96.96}&\textbf{93.62}&\textbf{94.26}\\
		{}&MFTRR(GCN)&{93.62}&{77.13}&{78.94} \\
		\bottomrule[2pt]
	\end{tabular}
	\caption{Multi-Level Evidence Relational Reasoning module that different structures.}
\end{table}

\subsection{Comparison of Experimental Results using Public Datasets }

\begin{table}[b!]
	\centering
	\begin{tabular}{  c | c | c  c  c }
		
		\hline
		\toprule[2pt]
		\textbf{Type} &\textbf{Method} &  \multicolumn{3}{c}{\textbf{The Lazada-Home}}     \\ 
		\textbf{} & \textbf{}& \textbf{MAP} &\textbf{N@3} &\textbf{N@5}    \\ \hline
		{Text-only} &       BiMPM&	70.6 &64.7& 69.1 \\
		{} &       Conv-KNRM&	71.40&	 65.70&	70.50 \\ \hline
		{Multimodal} & SSE-Cross&	72.20&	66.00&	71.00 \\
            {} & DIP& 73.89&	67.21&	72.16\\
		{} & MCR& 74.00&	67.80&	72.50 \\
		{} & The current method&	\textbf{74.71}&	\textbf{69.12}&	\textbf{73.95  } \\
		\bottomrule[2pt]
	\end{tabular}
	\caption{Topic-post relationship prediction results on the Lazada-Home dataset.}
\end{table}
To validate the effectiveness of the current model structure, the publicly available Lazada-MRHP dataset was chosen, which is similar to the multimodal post-topic dataset. This dataset was collected from the Lazada e-commerce platform, encompassing products and their corresponding user reviews. Two types of comparison models were used: one utilizing textual data and the other utilizing multimodal data. The comparison models for textual data included the Bilateral Multi-Perspective Matching (BiMPM) model and the Convolutional Kernel-based Neural Ranking Model (Conv-KNRM). For models using multimodal data, Stochastic Shared Embeddings Cross-attention (SSE-Cross), Multi-perspective Coherent Reasoning (MCR), and Dual Incongruity Perceiving (DIP) are chosen. The results of the comparison experiments, presented in Table 11, reveal a clear trend indicating that the model utilizing multimodal data outperforms the model that relies solely on textual data. Furthermore, the current model surpasses other models in performance, indicating that our model's architecture enables a more comprehensive analysis of the relationship between products and reviews through the integration of multimodal data.

\subsection{Ablation Study}
\begin{table}[b!]
	\centering
	\begin{tabular}{  c | c | c  c  c }
		
		\hline
		\toprule[2pt]
		\textbf{Datasets} &\textbf{Model Variant} &  \textbf{MAP}&  \textbf{N@3} & \textbf{N@5}   \\ \hline
		{The Education Course} & {MFTRR}& \textbf{89.06}&	\textbf{87.10}&	\textbf{88.98}    \\ 
		{}&	{-w/o multi-Level evidence}&	71.29&	45.27&	50.15 \\	
		{} & {-w/o multi-Level evidence I}&	88.25&	85.98&	86.68\\ 
		{} & {-w/o multi-Level evidence II}&	87.90&	 85.93&	86.94 \\
        {} & {-w/o local-global semantic}&	87.95&	85.88&	86.79 \\
		{} & {-w/o local-global semantic I}&	88.25&	85.36&	86.42 \\
		{} & {-w/o local-global semantic II}&	88.04&	86.04&	86.51 \\ \hline
		{The Excellence Courses}&	{MFTRR}&	\textbf{86.56}&	\textbf{85.07}&	\textbf{85.15} \\
		{}&	{-w/o multi-Level evidence}&	75.31&	 63.21&	65.93 \\
		{}&	{-w/o multi-Level evidence I}&86.28	&84.21&	84.72  \\
		{}&	{-w/o multi-Level evidence II}&86.01	&84.17&	84.69 \\
        {}&	{-w/o local-global semantic}&86.21&	84.33&	84.79 \\
		{}&	{-w/o local-global semantic I}&86.18&	84.38&	84.97 \\
		{}&	{-w/o local-global semantic II}&86.25&	84.37&	84.89 \\ \hline
		{The Art History Course}&	{MFTRR}&	\textbf{96.96}&	\textbf{93.62}&	\textbf{94.26} \\
		{}&	{-w/o multi-Level evidence}&	94.26&	77.62&	79.51 \\
		{}&	{-w/o multi-Level evidence I}&96.35&	92.15&	92.89 \\
		{}&	{-w/o multi-Level evidence II}&96.55&	92.51&	93.15 \\
        {}&	{-w/o local-global semantic}&96.59&	91.90&	92.92 \\
		{}&	{-w/o local-global semantic I}&96.46&	92.56&	93.35 \\
		{}&	{-w/o local-global semantic II}&96.38&	92.10&	92.91 \\
		\bottomrule[2pt]
	\end{tabular}
	\caption{The ablation study on the Education Course, the Excellence Courses, and the Art History Course datasets.}
\end{table}
Detailed ablation experiments were conducted to analyze the effectiveness of the different components of the model. These experiments involved removing the multi-level evidential relational reasoning module (denoted as -w/o multi-Level evidence), removing the topic-post significant information relationship evidence reasoning within the multi-level evidential relational reasoning module (denoted as -w/o multi-Level evidence I), removing the topic-post internal logic relationship evidence reasoning within the multi-level evidential relational reasoning module (denoted as -w/o multi-Level evidence II), removing the local-global semantic correlation reasoning module (denoted as -w/o local-global semantic), removing the local semantic correlation within the local-global semantic correlation reasoning module (denoted as -w/o local-global semantic I), removing the global semantic correlation within the local-global semantic correlation reasoning module (denoted as -w/o local-global semantic II). 

From Table 12, it can be observed that the multi-level evidential relational reasoning module has the greatest impact on the model's performance. This suggests that reasoning about the relationship between posts and topics from multiple levels can more fully evaluate the quality of posts. Within the multi-level evidential relational reasoning module, the numerical difference between the topic-post important information relationship evidence graph and the topic-post logical relationship graph is not significant, indicating that both components play important roles in the model's performance. Upon analyzing the data, it is evident that logical relationship evidence reasoning has a slightly greater influence than the topic-post significant relationship evidence reasoning. This may be due to the more granular nature of logical relationship evidence analysis compared to important relationship evidence analysis in post-quality evaluation. In the local-global semantic reasoning module, it is found that global semantic relationships have a greater impact on the model than local semantic relationships. This could be attributed to its ability to analyze posts and topics holistically, thereby better identifying the primary semantic relationship information between them rather than focusing on fragmented aspects.

\begin{figure}[htbp]
	\centering
	\begin{tabular}{  c  }
		\toprule[2pt]
		\large\makecell[l]{\textbf{Discussion Topics:}}   \\ 
		\makecell[l]{\small讨论题：饿了么在地铁站写了19句拼贴诗送给上海。请回答以下问题。(1)你认为} \\ 
       \makecell[l]{\small拼贴诗和饿了么的关联是什么？饿了么可以实现什么营销目的？(2)请举出一例成}\\
       \makecell[l]{\small功的地铁广告，并说明优点。}\\
       
		\begin{minipage}[b]{0.25\columnwidth}
			\centering
			\raisebox{-.5\height}{\includegraphics[width=\linewidth]{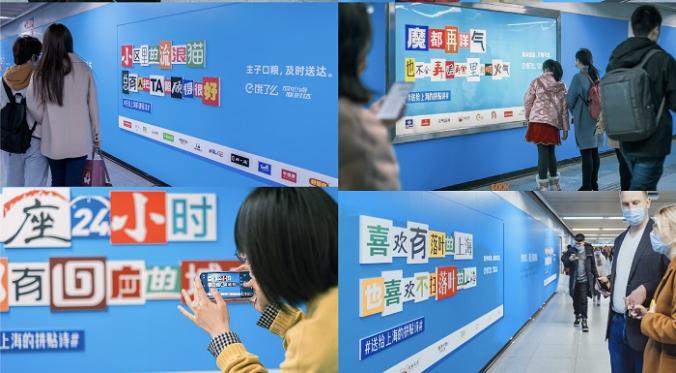}}
		\end{minipage}
		\\ \hline
		\large\makecell[l]{\textbf{post1:}}   \\
		\makecell[l]{\small1.关联：(1)拼贴诗中的字由饿了么旗下店名组成，温暖的诗句引发群众好感，引}  \\
        \makecell[l]{\small起对平台的关注。目的：(1)地铁人流量大提高知名度；(2)拼贴诗拉近客户的联系}  \\
        \makecell[l]{\small；(3)拼贴诗与上海人共情，增强对平台的信任。2.地铁广告：百度广告，广告词}  \\
        \makecell[l]{\small表达了老人的心声，既感动又达到宣传的效果，同时又建议每个儿女陪伴老人。}  \\
       
	   \begin{minipage}[b]{0.27\columnwidth}
	   	\centering
	   	\raisebox{-.5\height}{\includegraphics[width=\linewidth]{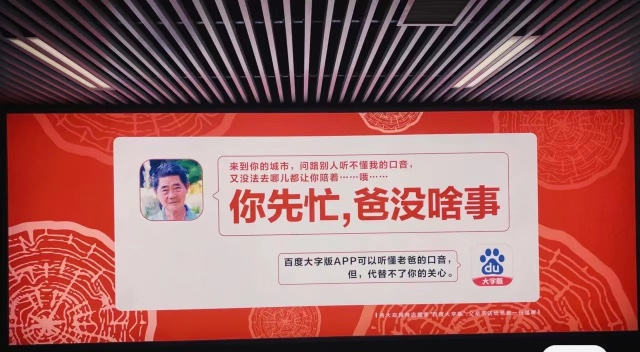}}
	   \end{minipage}
		\\ \hline
		\large\makecell[l]{\textbf{post2:}}   \\
        \makecell[l]{\small喜茶和蚂蚁森林的联名广告，色调和醒目logo给人眼前一亮。蚂蚁森林倡导绿色}  \\
        \makecell[l]{\small公益，喜茶与其合作带来了绿色生活的理念，也提高了品牌的影响力。}  \\
		 \begin{minipage}[b]{0.27\columnwidth}
			\centering
			\raisebox{-.5\height}{\includegraphics[width=\linewidth]{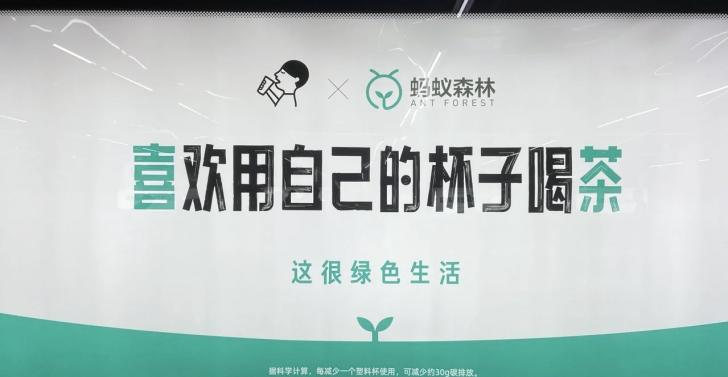}}
		\end{minipage}
		\\
		\bottomrule[2pt]
	\end{tabular}
	\caption{An example of discussion topic and posts. (\textbf{Top}) discussed topic: " ele.me wrote 19 collage poems for Shanghai at the metro station. Students answer the following questions. (1) What do you think is the connection between the collage poems and ele.me? What marketing purpose can achieve? (2) Give an example of a successful metro advert and explain its advantages. "; (\textbf{Middle}) post1: "1. Relevance: (1) The collage poem is composed of the shops' names in the platform, and the warm poem triggers the crowd's goodwill and attention. Purpose: (1) High metro foot traffic thus increasing the platform's visibility; (2) Bringing customers closer through the collage poem; (3) Empathising with Shanghainese to enhance the platform's trust. 2. Metro advertisement: The advertisement of Baidu, the slogan speaks to the heart of the elderly, which achieves the effect of publicity, and sincerely suggests every child accompany the elderly." ; (\textbf{Bottom}) post: “The co-branded advert between HEYTEA and Ant Forest is eye-catching with its hue and eye-catching logo. Ant Forest advocates green public welfare. The collaboration between HEYTEA  and Ant Forest brings the idea of green living, and improves the brand's influence”.}
  \clearpage 
\end{figure}

\subsection{Case Study}

To further illustrate the importance of analyzing multimodal topic-post in a multidimensional and multilevel way, a typical case from the dataset has been selected to demonstrate that the current model assesses post quality with greater accuracy. In the example presented in Figure 12, different models give different scores. Notably, models that rely solely on textual information give incorrect scores due to incomplete data analysis. Models that utilize multimodal data also give different results. The figure shows that post 2 fails to answer both questions thoroughly, answering only the second question. A single-dimensional analysis using multimodal data results in inaccuracies, as it solely analyses the correlation between the post and the topic without assessing whether the post adequately answers all questions. Therefore, evaluating post quality necessitates a multidimensional and multilevel analysis of the relationship between posts and discussion topics.

\section{Conclsion}

Accurately and comprehensively assessing the quality of posts is crucial for enabling teachers to gain timely insights into overall student learning effectiveness and to provide personalized interventions for individual students. In this work, a novel multimodal fine-grained topic-post relationship inference framework is introduced for the precise and comprehensive evaluation of posts. The framework leverages multimodal data to reconstruct the cognitive processes of human thinking across multi-scales and multi-levels. It comprises the local-global semantic relationship reasoning module and the multi-level evidence relationship reasoning module. The local-global semantic relationship reasoning module analyzes the semantic relationship between posts and topics through both local and global scales and filters the noise through the topic-based global fusion mechanism to obtain the maximum semantic relationship information based on the topic. The multi-level evidence relationship reasoning module captures the subtle relationships between a topic and a post through fine-grained topic-post relationship reasoning at different levels. The topic-post significant information evidence reasoning employs a retrieval attention mechanism that draws from the post-topic overall relational evidence graph and the post-topic representation to extract significant information oriented towards the relationship graph at a macro level. The topic-post internal logic relationship evidence reasoning constructs a relationship evidence graph by combining the overall internal relationship of the topic with the post to obtain the subtle relationship features between the topic and the post at a micro level. In addition, three multimodal topic-post datasets were constructed to evaluate the effectiveness of MFTRR. Experiments conducted on both the newly collected dataset and the Lazada-Home dataset demonstrate that MFTRR surpasses other baseline methods. Notably, on the Art History course dataset, it outperforms the leading text-only method by 9.52$\%$ in terms of the NDCG@3 metric.

\section{Future Works}

For the research direction of post-quality assessment, the future work and challenges are analyzed from the following aspects: (1) Effectively addressing the issue of missing multimodal data. Current models rely on high-quality multimodal data, and the absence of such  data significantly affects model performance. Therefore, models that can effectively generate missing modality data based on existing data or algorithms that can handle incomplete multimodal data can significantly enhance the effectiveness of post quality assessment. This is a technical challenge that future research needs to focus on overcoming. (2) Constructing a post quality assessment and explanation generation model. Current research only assesses post quality without providing explanations or justifications, which results in insufficient model interpretability and transparency. In the future, we plan to design a multimodal post quality assessment and explanation generation model that not only provides assessment scores but also offers relevant explanations to enhance learners' trust and understanding of the model's results, thereby better guiding and promoting students through evaluation. (3) Building a dataset for multimodal post quality assessment and explanation. High-quality datasets are crucial for advancing research in this field. To support the development and validation of multimodal post quality assessment and explanation models, we plan to specifically construct a high-quality dataset in the future, containing rich multimodal data and corresponding assessment explanation. (4)Developing a multimodal post quality assessment model based on large language models (LLMs). Recent research progress has shown that LLMs perform well in various fields. Our exploration will focus on fine-tuning pre-trained LLMs using multimodal data to construct a model specifically designed for assessment and explanation in the education field, further improving the accuracy and interpretability of post quality assessment.(5)Incorporating emotional modality information. Students' emotional tendencies can reflect their learning attitudes, and negative emotions may hinder the comprehensive and accurate completion of learning tasks. Therefore, emotional information can greatly assist in the assessment of post quality. Future research needs to delve deeper into how to effectively integrate emotional modality into post quality assessment models to improve the accuracy and comprehensiveness of the assessment. (6)Developing an evaluation and recommendation framework. This framework will analyze and evaluate post quality, identify students' knowledge weaknesses, and recommend relevant learning content, thereby facilitating personalized learning. (7) Incorporating data from other modalities to explore issues in the education field. The current framework has been applied in discussion board scenarios, utilizing multimodal data (text and images) to analyze the relationship between topics and posts. However, other scenarios within the education field may involve additional modalities, such as video. Future research needs to further explore how to effectively incorporate these additional modalities to more comprehensively analyze various issues in the education field.

\bibliographystyle{elsarticle-num} 
\bibliography{reference}





\end{CJK}
\end{document}